\documentclass[nohyperref]{article}

\usepackage{microtype}
\usepackage{graphicx}
\usepackage{subfigure}
\usepackage{booktabs} 

\usepackage{hyperref}

\usepackage[accepted]{icml2022}

\usepackage{amsmath}
\usepackage{amssymb}
\usepackage{bm}
\usepackage{mathtools}
\usepackage{amsthm}

\usepackage{microtype}
\usepackage{adjustbox}
\usepackage{multirow}
\usepackage[normalem]{ulem}
\useunder{\uline}{\ul}{}
\usepackage{bbding}
\usepackage{pifont}
\usepackage{wasysym}
\usepackage[capitalize,noabbrev]{cleveref}

\newcommand{\mr}[2]{\multirow{#1}{*}{#2}}

\usepackage{amsfonts}       
\usepackage{nicefrac}       
\usepackage{microtype}      
\usepackage{array}
\newcolumntype{P}[1]{>{\centering\arraybackslash}p{#1}}

\theoremstyle{plain}

\theoremstyle{definition}

\theoremstyle{remark}

\usepackage{enumitem}

\usepackage[textsize=tiny]{todonotes}

\DeclareMathOperator*{\concat}{\scalebox{1}[2.0]{$\parallel$}}

\icmltitlerunning{Local Augmentation for Graph Neural Networks}

\begin{document}

\twocolumn[
\icmltitle{Local Augmentation for
Graph Neural Networks}



\icmlsetsymbol{equal}{*}
\icmlsetsymbol{workdone}{\dag}

\begin{icmlauthorlist}
\icmlauthor{Songtao Liu}{psu,workdone}
\icmlauthor{Rex Ying}{stanford}
\icmlauthor{Hanze Dong}{hkust}
\icmlauthor{Lanqing Li}{tencent}
\icmlauthor{Tingyang Xu}{tencent}
\icmlauthor{Yu Rong}{tencent}
\icmlauthor{Peilin Zhao}{tencent}
\icmlauthor{Junzhou Huang}{tencent}
\icmlauthor{Dinghao Wu}{psu}
\end{icmlauthorlist}

\icmlaffiliation{psu}{The Pennsylvania State University}
\icmlaffiliation{stanford}{Stanford University}
\icmlaffiliation{hkust}{Hong Kong University of Science and Technology}
\icmlaffiliation{tencent}{Tencent AI Lab}

\icmlcorrespondingauthor{Songtao Liu}{skl5761@psu.edu}
\icmlcorrespondingauthor{Rex Ying}{rexying@stanford.edu}
\icmlcorrespondingauthor{Lanqing Li}{lanqingli1993@gmail.com}
\icmlcorrespondingauthor{Dinghao Wu}{dinghao@psu.edu}

\icmlkeywords{Machine Learning, ICML}

\vskip 0.3in
]



\printAffiliationsAndNotice{\textsuperscript{\dag}Part of the work was done during Songtao's internship at Tencent AI Lab}

\begin{abstract}
Graph Neural Networks (GNNs) have achieved remarkable performance on graph-based tasks. The key idea for GNNs is to obtain informative representation through aggregating information from local neighborhoods. However, it remains an open question whether the neighborhood information is adequately aggregated for learning representations of nodes with few neighbors. To address this, we propose a simple and efficient data augmentation strategy, local augmentation, to learn the distribution of the node features of the neighbors conditioned on the central node's feature and enhance GNN's expressive power with generated features. Local augmentation is a general framework that can be applied to any GNN model in a plug-and-play manner. It samples feature vectors associated with each node from the learned conditional distribution as additional input for the backbone model at each training iteration. Extensive experiments and analyses show that local augmentation consistently yields performance improvement when applied to various GNN architectures across a diverse set of benchmarks. For example, experiments show that plugging in local augmentation to GCN and GAT improves by an average of 3.4\% and 1.6\% in terms of test accuracy on Cora, Citeseer, and Pubmed. Besides, our experimental results on large graphs (OGB) show that our model consistently improves performance over backbones. Code is available at \url{https://github.com/SongtaoLiu0823/LAGNN}.

\end{abstract}

\section{Introduction}
Graph Neural Networks (GNNs) and their variants~\citep{kipf2017semi,hamilton2017inductive,velivckovic2018graph} have achieved state-of-the-art performance on a variety of graph-based tasks, including recommendation system~\citep{ying2018graph}, drug discovery~\citep{dai2019retrosynthesis} and traffic prediction~\citep{guo2019attention}. The core of GNNs is to employ a message-passing mechanism that passes and aggregates information from the local neighborhood to generate informative representations.

Recent development of deep GNNs, such as JKnet~\citep{xu2018representation}, GCNII~\citep{chen2020simple}, and RevGNN-Deep~\citep{pmlr-v139-li21o} adds the output of shallow layers to the deep layers with a residual-style design, to preserve the locality information of node representations~\citep{chen2020simple}. Moreover, recent studies~\citep{zeng2021decoupling, zhang2021nested, wijesinghe2022a} utilize structural information of the local neighborhood to design efficient message-passing aggregation schemes to enhance the expressive power of GNNs. These works demonstrate that local information plays a significant role in training GNN models and designing powerful GNNs.

Despite advances of GNNs in learning node representations from the local neighborhood, it remains an open problem whether the local neighborhood information is sufficient to obtain effective node representations, especially for nodes with \emph{limited} number of neighbors. 
We argue that the \emph{limited} number of neighbors in the local neighborhood restricts the expressive power of GNNs and hinders their performance, especially in sample-starving cases where some nodes have very few neighbors.
Stacking graph layers to enlarge the receptive field can incorporate multi-hop neighboring information but leads to over-smoothing~\citep{li2018deeper} without residual connection to the input, and is not a direct solution to address this issue. Existing works on GNN model architecture cannot tackle the problem that the very limited neighbors are unfavorable to learning node representations. Therefore, here we focus on enriching the local information for low-degree nodes to obtain effective representations.

One promising solution is to generate more features for the local neighborhood via data augmentation. Data augmentation has been well-studied in computer vision~\citep{shorten2019survey,cubuk2019autoaugment,zhao2019data,dong2022learning} and natural language processing~\citep{fadaee2017data,csahin2019data,xia-etal-2019-generalized}, but remains under-explored on graph-structured data. Existing graph data augmentation approaches only perturb at the topology-level and feature-level from a global perspective, which can be divided into two categories: topology-level augmentation~\citep{rong2020dropedge, zhao2021data} and feature-level augmentation~\citep{deng2019batch, feng2019graph, kong2020flag,fang2021mathematical}. Topology-level augmentation perturbs the adjacency matrix, yielding different graph structures. On the other hand, existing feature-level augmentation ~\citep{deng2019batch, feng2019graph, kong2020flag} generate perturbations on node attributes with adversarial training to boost generalization. These augmentation techniques have a prominent drawback: they focus on global augmentation concerning the properties of the whole distribution of the graph rather than a single node, and neglect the local information of the neighborhood.

\begin{figure*}[t]
\centering
\includegraphics[width=16cm]{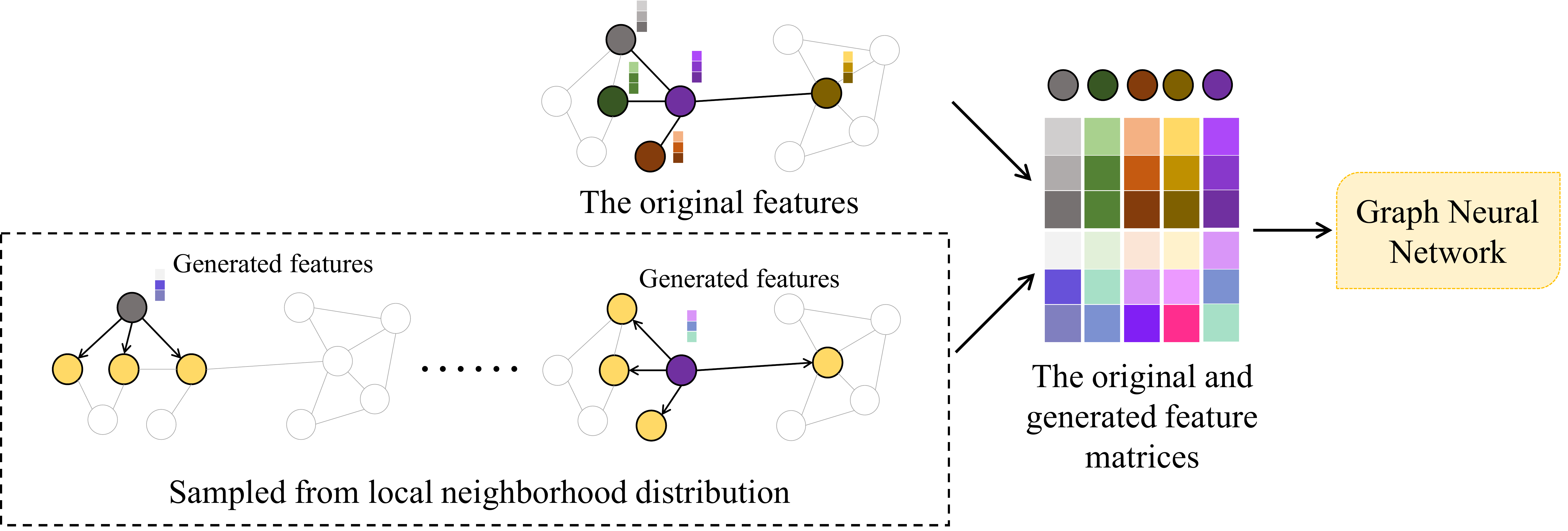}
\caption{A schematic depiction of our local augmentation. The yellow circles on the graph correspond to the neighbor nodes. Assume we have learned the distribution of the local neighborhood. We generate features from the local neighborhood distribution. And then we take the original features and the generated features as input for downstream GNNs.}
\label{fig:framework}
\end{figure*}

In this work, in order to promote the aggregation scheme with more generated features in the local neighborhood, we propose a novel and efficient data augmentation framework: \textbf{L}ocal \textbf{A}ugmentation for \textbf{G}raph \textbf{N}eural \textbf{N}etworks (LA-GNNs). The term ``local augmentation'' refers to generating neighborhood features via a generative model \textbf{conditioned on local structures and node features}. Specifically, our proposed framework includes a pre-training step, which learns the conditional distribution of the connected neighbors' node features given one center node's feature via a generative model. As shown in Fig.~\ref{fig:framework}, we then exploit this distribution to generate feature vectors associated with this center node as additional input at each training iteration. Furthermore, we decouple the pre-training of the generative model and downstream GNN training, allowing our data augmentation model to be applied to any GNN model in a plug-and-play manner.

We verify the effectiveness of LAGNNs on three standard citation networks (Cora, Citeseer, Pubmed) and Open Graph Benchmark (OGB)~\citep{hu2020open}. Extensive experimental results on semi-supervised node classification show that our local augmentation achieves new stat-of-the-art performance: LAGCN and LAGAT achieve up to by an average of 3.4\% and 1.6\% in terms of test accuracy over GCN and GAT respectively on Cora, Citeseer, and Pubmed. LAGNN also obtains superior performance on large-scale OGB datasets. We show that our model improves 1.7\% and 0.2\% of test accuracy on Pubmed for nodes with degrees in $[2, 5]$ and $[6, 20]$ respectively. Besides, our local augmentation model outperforms other feature/topology-level augmentation models, such as G-GNN~\citep{zhu2020pre}, DropEdge~\citep{rong2020dropedge}, G\textsc{rand}~\citep{feng2020grand}, and G\textsc{aug}~\citep{zhao2021data} on semi-supervised node classification tasks, which demonstrates the superiority of our model. \paragraph{Contributions.} 
\begin{itemize}[leftmargin=10pt]
    \item We propose a general augmentation strategy to generate more features in the local neighborhood to enhance the expressive power of existing GNNs;
    \item We explore a new direction on pre-training generative models for graphs to improve downstream task performance;
    \item Our proposed framework is flexible and can be applied to various popular backbones. Extensive experimental results demonstrate that our proposed framework could improve the performance of GNN variants on different benchmark datasets.
\end{itemize}

\section{Preliminaries}
\paragraph{Notations.} Let $\mathcal{G}=(\mathcal{V}, \mathcal{E})$ represent the graph, where $\mathcal{V}$ is the set of vertices $\left\{v_{1}, \cdots, v_{N}\right\}$ with $|\mathcal{V}|=N$ and $\mathcal{E}$ is the set of edges. The adjacency matrix is defined as $\bm{A} \in \{0,1\}^{N \times N}$, and $\bm{A}_{i,j}=1$ if and only if $\left(v_{i}, v_{j}\right) \in \mathcal{E}$. Let $\mathcal{N}_{i}=\{v_j | \bm{A}_{i,j} = 1\}$ denotes the neighborhood of node $v_i$ and $\bm{D}$ denote the diagonal degree matrix, where $\bm{D}_{i,i}=\sum_{j=1}^{n} \bm{A}_{i,j}$. The feature matrix is denoted as $\bm{X} \in \mathbb{R}^{N \times F}$ where each node $v$ is associated with a $F$-dimensional feature vector $\bm{X}_v$. $\bm{Y} \in \{0,1\}^{N \times C}$ denote the one-hot label matrix, where $\bm{Y}_i \in \{0, 1\}^{C}$ is a one-hot vector and $\sum_{j=1}^{C} \bm{Y}_{i,j}=1$ for any $v_i \in V$.
\paragraph{Graph Neural Networks.} 
Graph Neural Networks (GNNs) directly operate on the graph structure and aggregate information via message-passing mechanism~\citep{zhou2018graph}. They repeatedly aggregate the representations of immediate neighbors $\mathcal{N}_{v}$ of node $v$ and combine the aggregated information and its representation vector to obtain a representation vector $\bm{h}_v$. The $k$-th layer of the GNN message-passing scheme is:
\begin{equation}
\bm{h}_{v}^{(k)}=\mathsf{COM}\left(\bm{h}_{v}^{(k-1)}, \mathsf{AGG}\left(\left\{\bm{h}_{u}^{(k-1)}, \bm{e}_{u,v}|u\in\mathcal{N}_v\right\}\right)\right),
\end{equation}
where $\mathsf{COM}(\cdot)$ and $\mathsf{AGG}(\cdot)$ denotes COMBINE and AGGREGATE functions respectively, $\bm{h}_{v}^{(k)}$ is the representation vector of node $v$ in the $k$-th layer, and $\bm{e}_{u,v}$ is the edge vector between node $u$ and node $v$. Specifically, $\bm{h}_{v}^{(0)} = \bm{X}_v$.

\section{Local Augmentation for Graph Neural Networks (LAGNN)}
In this section, we first present how to generate more features in the local neighborhood via a generative model. Then we show how to decouple the pre-training of the generative model and downstream GNN training from a probabilistic perspective, so that our local augmentation model can be applied to any GNN model in a plug-and-play manner. We then introduce the architecture of LA-GNNs and the training details. The overall framework is illustrated in Fig.~\ref{fig:framework}.

\subsection{Local Augmentation}
\label{Local Augmentation}
\paragraph{Motivation.}Existing GNNs focus on designing a message-passing scheme to exploit local information to obtain node representations. We explore a new direction in that we can generate more features in the local neighborhood, especially for nodes with few neighbors, to enhance the expressive power of various GNNs. In order to generate more features in a node $v$'s neighborhood $\mathcal{N}_v$, we need to know the distribution of the node features of its neighbors. As this distribution is related to the center node $v$, we can learn it conditioned on the center node's feature via a generative model. 

\paragraph{Benefits.} Compared to training generative models for each node, there are three benefits to train a single generative model for all nodes 1) Learning the conditional distribution across all nodes on the graph via a generative model reduces the computational cost. 2) In the generation stage, we can apply a specific node's feature vector as input (condition) and generate feature vectors associated with this node. 3) It has better scalability and generalization. For a new node added to a dynamic graph, we can directly generate feature vectors without re-training a new generative model since the generative model contains such generalized information. So our local augmentation model can be applied to inductive learning tasks, such as graph classification.

\paragraph{Approach.}
We exploit the conditional variational auto-encoder (CVAE)~\citep{kingma2013auto, sohn2015learning} to learn the conditional distribution of the node features of connected neighbors $u (u \in \mathcal{N}_{v})$ given the center node $v$. In our CVAE setting, we use $\bm{X}_v$ as a condition since the distribution of $\bm{X}_u (u \in \mathcal{N}_{v})$ is related to $\bm{X}_v$. Following \citet{sohn2015learning}, the latent variable $\mathbf{z}$ is generated from the prior distribution $p_\theta(\mathbf{z}|\bm{X}_v)$ and the data $\bm{X}_u$ is generated by the generative distribution $p_\theta(\bm{X}|\bm{X}_v, \mathbf{z})$ conditioned on $\mathbf{z}$ and $\bm{X}_v$: $\mathbf{z} \sim p_\theta(\mathbf{z}|\bm{X}_v)$, $\bm{X}_u \sim p_\theta(\bm{X}|\bm{X}_v, \bm{z}_v)$. Let $\phi$ denote the variational parameters and $\theta$ represent the generative parameters, we have 
\begin{equation*}
\begin{split}
\log p_{\theta}\left(\bm{X}_{u} | \bm{X}_{v}\right) &= \int q_{\phi}(\mathbf{z} | \bm{X}_u, \bm{X}_v) \log \frac{p_{\theta}(\bm{X}_u, \mathbf{z} | \bm{X}_v)}{q_{\phi}(\mathbf{z} | \bm{X}_u, \bm{X}_v)} \mathrm{d} \mathbf{z}\\
&\quad +KL\left(q_{\phi}(\mathbf{z} | \bm{X}_u, \bm{X}_v) \| p_{\theta}(\mathbf{z} | \bm{X}_{u}, \bm{X}_v)\right) \\ &\geq \int q_{\phi}(\mathbf{z} | \bm{X}_u, \bm{X}_{v}) \log \frac{p_{\theta}(\bm{X}_u, \mathbf{z} | \bm{X}_v)}{q_{\phi}(\mathbf{z} | \bm{X}_u, \bm{X}_v)} \mathrm{d} \mathbf{z},
\end{split}
\end{equation*}
and the evidence lower bound (ELBO) can be written as:
\begin{equation}
\label{generator elbo}
\begin{split}
\mathcal{L}(\bm{X}_u, \bm{X}_v ; \theta, \phi) =& -KL(q_{\phi}(\mathbf{z} | \bm{X}_u, \bm{X}_v) \| p_{\theta}(\mathbf{z} | \bm{X}_v))\\
&+\frac{1}{L}\sum_{l=1}^L\log p_\theta(\bm{X}_u|\bm{X}_v,\mathbf{z}^{(l)})
\end{split}
\end{equation}
where $\mathbf{z}^{(l)}=g_\phi(\bm{X}_v, \bm{X}_u, \epsilon^{(l)})$, $\epsilon^{(l)} \sim \mathcal{N}(\mathbf{0}, \mathbf{I})$ and $L$ is the number of neighbors of node  $v$. Note that as we have discussed before, we just train one CVAE for all nodes. In the training stage, the objective is to use the neighboring pairs ($\bm{X}_v$, $\bm{X}_u$, $u\in \mathcal{N}_v$) as input to maximize the ELBO, i.e., Eq. (\ref{generator elbo}). In the generation stage, we use the node feature $\bm{X}_v$ as the condition and sample a latent variable $\mathbf{z} \sim \mathcal{N}(\mathbf{0}, \mathbf{I})$ as input for the decoder. Then we can get generated feature vector $\bm{\overline{X}}_v$ associated with node $v$.
\paragraph{Discussion.}
When learning the distribution of the neighbors' features conditioned on the central node, we do not consider the effect of other nodes connected to each neighbor on the neighbors' features. If we regard the center node as a parent node, and its neighbors as the children nodes of the center node, then our assumption is similar to the Causal Markov Condition in a Bayesian network~\citep{hausman1999independence}: the distribution of the neighbor’s features is independent of its non-descendants given its parent node. This assumption is important and common in the literature of the probabilistic graphical model. The advantage is that this assumption avoids the exponential complexity of conditioning on multi-hop neighbors, significantly improving scalability. Our experimental results show that our method still achieves remarkable performance across all benchmarks, thanks to the expressive power of deep generative models (similar to how the assumption of variational inference does not limit the performance of deep VAEs in real datasets).

\subsection{Decoupling the Generative Model Training from Downstream Graph Learning}
Most existing GNN models follow the message passing mechanism~\citep{gilmer2017neural} and can be regarded as a learned classification or regression function. In order to make predictions, GNN models need to estimate the posterior distribution $P_\Theta(\cdot|\bm{A},\bm{X})$ with respect to the graph structure $\bm{A}$ and feature matrix $\bm{X}$. For example, $\cdot$ can be class labels $\bm{Y}$ on the node classification task. We can use Maximum Likelihood Estimation (MLE) to estimate the parameter $\Theta$ by optimizing the following likelihood function:
\begin{equation}
\label{likelihood function}
\max \prod_{i} P_\Theta\left(\cdot | \bm{A}, \bm{X}\right),
\end{equation}
where $i$ represents the $i$-th data point in the training dataset. In our local augmentation model, in order to further improve the expressive power of GNNs, we introduce a generated feature vector $\bm{\overline{X}}_v$ for the center node $v$ by using $\bm{X}_v$ as input condition and sampling from the generative model. Let $\bm{\overline{X}}$ denote the generated feature matrix where the $j$-th row corresponds to the generated feature vector $\bm{\overline{X}}_j$. We incorporate $\bm{\overline{X}}$ in Eq. (\ref{likelihood function}) and rewrite it as follows:
\begin{equation}
\label{generated x likelihood}
\max \prod_{i}\int_{\bm{\overline{X}}} P_\Theta\left(\cdot, \bm{\overline{X}}| \bm{A}, \bm{X}\right).
\end{equation}
For Bayesian tractability, we decompose $P_\Theta$ in Eq.(\ref{generated x likelihood}) as a product of two posterior probabilities:
\begin{equation}
\label{two terms of distribution}
P_{\Theta, \Phi}(\cdot, \bm{\overline{X}}|\bm{A}, \bm{X}) := P_{\Theta}(\cdot|\bm{A}, \bm{X}, \bm{\overline{X}})Q_{\Phi}(\bm{\overline{X}}|\bm{A}, \bm{X}),
\end{equation}
where $P_{\Theta}(\cdot|\bm{A}, \bm{X}, \bm{\overline{X}})$ and $Q_{\Phi}(\bm{\overline{X}}|\bm{A}, \bm{X})$ denote the probabilistic distributions approximated by the GNN models and the generative model respectively, parameterized by $\Theta$ and $\Phi$. By doing this, we can decouple our proposed local augmentation and the specific graph learning, allowing our augmentation model to be applied to various GNN models with only one-time pre-training for the generative model. Therefore, local augmentation can be regarded as an unsupervised pre-training model prior to the GNN training. The representation power of Eq. (\ref{two terms of distribution}) is superior than that of a single predictor $P_\Theta\left(\bm{Y}_{k} | \bm{A}, \bm{X}\right)$ since we provide GNN models with more generated features in the local neighborhood.
\subsection{Architecture}
\label{architecture}
In this section, we present the details of how to train GNNs with generated features from our local augmentation model as additional input. To illustrate the effectiveness of our local augmentation model, we provide two different ways of exploiting our generated features, leading to average and concatenation design of the architectures.
\paragraph{LAGCN.}
\label{lagcn}
For GCN, we only make a small change on the first graph convolution layer with
\begin{equation}
    \bm{H}^{(1)} = \sigma\left( \widetilde{\bm{P}}\bm{X}\bm{W}_0^{(1)}\right) \concat \sigma\left( \widetilde{\bm{P}}\bm{\overline{X}}\bm{W}_1^{(1)}\right),
\end{equation}
where $\widetilde{\bm{P}}=\widetilde{\bm{D}}^{-\frac{1}{2}} \widetilde{\bm{A}} \widetilde{\bm{D}}^{-\frac{1}{2}}$. The notation $\concat$ means concatenation of the matrices on the second dimension. The sub- and super- scripts in the weight matrix $\bm{W}$ denote the layers’ and the parameters’ ordinal numbers. In order not to change the parameter size of the GCN model, the sum of the second dimension of $\bm{W}_0^{(1)}$ and $\bm{W}_1^{(1)}$ is equal to the second dimension of $\bm{W}^{(1)}$ of GCN. For other architectures (LAGAT, LASAGE, LAGCNII) which we will discuss later, we keep the same setting on the parameter size in the first layer as LAGCN. For GraphSAGE and GCNII, they have similar architectures to GCN and we employ the same modification strategy as LAGCN for LASAGE and LAGCNII. In addition to the concatenation-style design, we can also average $\bm{X}$ and $\bm{\overline{X}}$ as input for GNNs and do not change the architecture.

\paragraph{LAGAT.}
Similarly, the first layer of LAGAT is defined as follows:
\begin{equation}
\begin{split}
\bm{H}^{(1)}=& \left(\concat_{d=1}^{D/2} \sigma\left(\sum_{u \in \mathcal{N}_{v}} \alpha_{v u}^{d} \mathbf{W}_d^{(1)} \bm{X}_{u}\right)\right) \\ &\bigg{\|} \left(\concat_{d=D/2+1}^{D} \sigma\left(\sum_{u \in \mathcal{N}_{v}} \alpha_{v u}^{d} \mathbf{W}_d^{(1)} \bm{\overline{X}}_{u}\right) \right),
\end{split}
\end{equation}
where $\alpha_{vu}^d$ is computed on $\bm{X}$ ($1\leq d \leq D/2$) or $\bm{\overline{X}}$ ($D/2+1\leq d \leq D$). Note that the second dimension of $\mathbf{W}_d^{(1)}$ is the same as GAT. We just replace the input of the half of the attention heads with $\bm{\overline{X}}$. And we also provide the average-style design for GAT.

\begin{algorithm}[t]
\caption{Local Augmentation for Graph Neural Networks} 
\label{training framework}
\begin{algorithmic}[1]
\STATE {\bfseries Input:} Adjacency matrix $\bm{A}$, feature matrix $\bm{X}$ 
\STATE {\bfseries Output:} Prediction $\bm{Z}$
\STATE Pre-train the generative model $Q_{\Phi}$ using Eq. (\ref{generator elbo}), given $\bm{A}$ and $\bm{X}$ as input.
\WHILE{not convergence}
\FOR{$k=1:K$} 
\STATE Generate the augmented feature matrix:  
${\bm{\overline{X}}}^{(k)} \sim Q_{\Phi}$. 
\STATE Obtain the prediction using LAGNN $P_{\Theta}$:
${\bm{Z}}^{(k)} = f_{LAGNN}(\bm{A}, \bm{X}, \bm{\overline{X}}^{(k)}, \Theta)$
\ENDFOR
\STATE Compute supervised classification loss $\mathcal{L}_{s}$ via Eq. (\ref{supervised loss}) 
\STATE Optionally compute the consistency regularization loss $\mathcal{L}_{c}$ via Eq. (\ref{consistency loss}).
\STATE Update the parameters $\Theta$ via gradient descent:
$\Theta = \Theta - \eta \nabla_\Theta (\mathcal{L}_{s} (+ \beta \mathcal{L}_{c}))$
\STATE Regenerate the augmented feature matrix:  
${\bm{\overline{X}}} \sim Q_{\Phi}$. 
\STATE Compute the validation loss function or the validation accuracy via ${\bm{Z}} = f_{LAGNN}(\bm{A}, \bm{X}, \bm{\overline{X}}, \Theta)$ and Eq. (\ref{supervised loss}).
\ENDWHILE
\STATE Predict via: ${\bm{Z}} = f_{LAGNN}(\bm{A}, \bm{X}, \bm{\overline{X}}, \Theta)$, where we select $\bm{\overline{X}}$ with the smallest validation loss function or the highest validation accuracy.
\end{algorithmic}
\end{algorithm}
\subsection{Loss Functions}
In this section, we explain two loss functions to train LAGNNs and the underlying motivation:
\emph{supervised loss} and \emph{consistency loss}.

\paragraph{Supervised Loss.}
We use the original feature matrix $\bm{X}$ and the generated feature matrix $\bm{\overline{X}}$ as input to train LAGNNs. Given the training labels $\bm{L}_T$ and $K$ augmented feature matrices $\bm{\overline{X}}^{(k)}$, we can write the supervised loss function for node classification tasks as follows: 
\begin{equation}
\label{supervised loss}
\mathcal{L}_{s} = -\frac{1}{K}\sum_{k=1}^K\sum_{i \in \bm{L}_T} \bm{Y}_{i} \log \bm{Z}_{i}^{(k)},
\end{equation}
where $\bm{Z}^{(k)}=f_{LAGNN}(\bm{A}, \bm{X}, \bm{\overline{X}}^{(k)}, \Theta)$. Note that we just provide a type of supervised loss function. For other graph learning tasks such as link prediction and graph classification, the supervised loss functions can be adjusted accordingly. 

\paragraph{Consistency Regularization Loss.}
Inspired by the huge success of consistency training~\citep{wang2020nodeaug, feng2020grand, sajjadi2016regularization, samuli2017temporal, berthelot2019mixmatch, verma2019interpolation} on semi-supervised learning tasks, we provide an optional loss function for specific GNNs and graph learning tasks. Intuitively, the consistency regularization encourages invariant prediction of different inputs at each training iteration~\citep{verma2019interpolation}. Specifically, we use the consistency regularization loss function proposed by \citet{feng2020grand} with the following form:
\begin{equation}
\label{consistency loss}
        \mathcal{L}_{c} = \frac{1}{K} \sum_{k=1}^{K} \sum_{i=1}^{N}\left\|\bm{\overline{Z}}_{i}^{\prime}-\bm{Z}_{i}^{(k)}\right\|_{2}^{2},
\end{equation}
where $\bm{\overline{Z}}_{i}= \frac{1}{K}\sum_{k=1}^K\bm{Z}_i^{(k)}$, $\bm{\overline{Z}}_{i}^{\prime} = \bm{\overline{Z}}_{i}^{\frac{1}{T}} / \sum_{c=1}^{C} \bm{Z}_{ic}^{\frac{1}{T}}$ is the sharpening trick~\citep{berthelot2019mixmatch}, and $T$ is a hyper-parameter that adjusts the ``temperature" of this categorical distribution. The sharpening trick can reduce the entropy of the predictions.

\paragraph{Training and Inference.}
The details of our training and inference process are outlined in Algorithm~\ref{training framework}. First, we train CVAE $i.e.$ our local augmentation model. And then we sample a different feature matrix generated by CVAE as additional input at each training iteration to train the GNN models. But for G\textsc{rand}~\citep{feng2020grand}, we just sample one feature matrix during training stage since we find we can get better performance with such a sample strategy. Supervised loss functions are computed on the initial feature matrix $\bm{X}$ and the generated feature matrix $\bm{\overline{X}}$. Besides, we optionally compute the consistency regularization loss function $\mathcal{L}_{c}$ based on $f_{LAGNN}(\bm{A}, \bm{X}, \bm{\overline{X}}^{(k)}, \Theta)$ for specific tasks. Moreover, we resample another feature matrix that is different from the one to compute the training loss function to compute the validation loss function and validation accuracy at each training iteration. At the inference stage, we don't need to generate $\bm{\overline{X}}$ again since we select $\bm{\overline{X}}$ with the smallest validation loss function on citation datasets (Cora, Citeseer, Pubmed) or the highest validation accuracy on OGB datasets. The computational complexity of pre-training and GNN training is $\mathcal{O}(|\mathcal{E}|(FD+LD^2))$ and $\mathcal{O}(S|\mathcal{E}|(FD+LD^2))$ respectively, where $D$ is the number of hidden channels and $L$ is the number of layers. The number of pre-training epochs is usually fewer than 10, which demonstrates that pre-training introduces little computational overhead.

\section{Experiments}
In this section, we evaluate the performance of our local augmentation model on various tasks including node classification, link prediction, and graph classification. All the experiments are conducted on open graph datasets.
\begin{table}[t]
\setlength{\tabcolsep}{1.7mm}
	\caption{Classification results on three citation networks (\%)}
	\label{fixed split}
	\vskip 0.15in
	\centering
	\begin{tabular}{lccc}
		\toprule 
		\textbf{Method} & \textbf{Cora} & \textbf{Citeseer} & \textbf{Pubmed} \\
		\midrule 
        Chebyshev        &81.2  &69.8  &74.4   \\
        APPNP     &83.8{$\pm$0.3}  &71.6{$\pm$0.5}  &79.7{$\pm$0.3}   \\
        MixHop     &81.9{$\pm$0.4}  &71.4{$\pm$0.8}  &80.8{$\pm$0.6}   \\
        Graph U-net    &84.4{$\pm$0.6}  &73.2{$\pm$0.5}  &79.6{$\pm$0.2}   \\
        GSNN-M      &83.9{$\pm$0.5}  &72.2{$\pm$0.5}  &79.1{$\pm$0.3}   \\
        S$^{2}$GC      &83.5{$\pm$0.02}  &73.6{$\pm$0.09}  &80.2{$\pm$0.02}   \\
        \midrule
        GCN     &81.5{$\pm$0.5}  &70.3{$\pm$0.7}  &79.0{$\pm$0.5}   \\
        G-GCN  &83.7  &71.3  &80.9\\
        DropEdge-GCN   &82.8  &72.3  &79.6\\
        \textsc{Gaug-O}-GCN    &83.6{$\pm$0.5}  &73.3{$\pm$1.1}  &79.3{$\pm$0.4}\\
        GraphSNN$_{GCN}$ & 83.1{$\pm$1.8} & 72.3{$\pm$1.5} & 79.8{$\pm$1.2} \\
        G\textsc{rand}-GCN   &84.5{$\pm$0.3}  &74.2{$\pm$0.3}  &80.0{$\pm$0.3}   \\
        LA-GCN    &\textbf{84.6{$\pm$0.5}}  &\textbf{74.7{$\pm$0.5}}  & \textbf{81.7{$\pm$0.7}}\\
        \midrule
        GAT        &83.0{$\pm$0.7}  &72.5{$\pm$0.7}  &79.0{$\pm$0.3}   \\
        \textsc{Gaug-O}-GAT &82.2{$\pm$0.2} &71.6{$\pm$1.1} & OOM \\
        GraphSNN$_{GAT}$ & 83.8{$\pm$1.2} & 73.5{$\pm$1.6} & 79.6{$\pm$1.4} \\
        G\textsc{rand}-GAT   &84.3{$\pm$0.4}  &73.2{$\pm$0.4}  &79.2{$\pm$0.6}   \\
        LA-GAT      &\textbf{84.7{$\pm$0.4}}  &\textbf{73.7{$\pm$0.5}} &\textbf{81.0$\pm$0.4} \\
        \midrule
        GCNII     &85.5{$\pm$0.5}  &73.4{$\pm$0.6}  &80.2{$\pm$0.4}   \\
        LA-GCNII    &\textbf{85.7{$\pm$0.3}}  &\textbf{74.1{$\pm$0.5}}  &\textbf{80.6{$\pm$0.7}} \\
         \midrule
        G\textsc{rand}   &85.4{$\pm$0.4}  &75.4{$\pm$0.4}  &82.7{$\pm$0.6}   \\
        \textsc{LA-}G\textsc{rand}      &\textbf{85.7{$\pm$0.3}}  &\textbf{75.8{$\pm$0.5}}  &\textbf{83.4{$\pm$0.6}}   \\
        \bottomrule
	\end{tabular}
\end{table}

\subsection{Semi-supervised Learning}
\paragraph{Datasets.}
We utilize three public citation network datasets Cora, Citeseer, and Pubmed~\citep{sen2008collective} for semi-supervised node classification. All the dataset statistics can be found in Appendix~\ref{appendix:reproducibility}.
\paragraph{Baselines.}
We consider three popular graph neural networks: GCN~\citep{kipf2017semi}, GAT~\citep{velivckovic2018graph}, and GCNII~\citep{chen2020simple} as our implemented backbones. GCN and GAT are representatives of the state-of-the-art GNN architectures, while GCNII is a deep GNN model with a skip connection design. For each of these backbones, we employ the concatenation-style design discussed in Sec.~\ref{architecture} as our LAGNN architecture. But we keep the size of the learnable weight matrix of LAGNN the same as the corresponding GNN model, which is detailed in the Appendix~\ref{appendix:reproducibility}. We also combine our method with other data augmentation models - G\textsc{rand}~\citep{feng2020grand}. To evaluate our proposed framework, we compare our model against state-of-the-art models of four categories: 
\begin{itemize}[leftmargin=10pt]
    \item Backbone models: Chebyshev~\citep{defferrard2016convolutional}, GCN~\citep{kipf2017semi}, GAT~\citep{velivckovic2018graph}, APPNP~\citep{klicpera2019predict}, Graph U-net~\citep{gao2019graph}, MixHop~\citep{abu2019mixhop}, GCNII~\citep{chen2020simple}, GSNN-M~\citep{wang2020graph}, and S$^{2}$GC~\citep{zhu2021simple}
    \item Feature-level augmentation models: G-GNNs~\citep{zhu2020pre}, and G\textsc{rand}~\citep{feng2020grand}.
    \item Topology-level augmentation modes: DropEdge~\citep{rong2020dropedge} and \textsc{Gaug-O}~\citep{zhao2021data}.
    \item Subgraph GNN: GraphSNN~\citep{wijesinghe2022a}. 
\end{itemize}
The choice of baselines aims to show that existing GNNs benefit from our proposed local data augmentation, and our model outperforms other data augmentation models and subgraph GNNs.
\paragraph{Experimental setup.} 
We apply the standard fixed splits~\citep{yang2016revisiting} on Cora, Citeseer, and Pubmed, with 20 nodes per class for training, 500 nodes for validation, and 1,000 nodes for testing. See more details on experimental setup and hyper-parameters in the Appendix~\ref{appendix:reproducibility}.
\paragraph{Comparison with SOTA.}
We report the mean node classification accuracy after 100 runs in Table~\ref{fixed split}. We reuse the metrics of the baselines already reported in the corresponding papers. The results demonstrate that the backbone models equipped with our method achieve better performance across all three datasets. Specifically, local augmentation can improve GCN by 3.1\%, 4.4\%, and 2.7\% on Cora, Citeseer, and Pubmed respectively, while the improved performance of LAGAT over GAT are 1.7\%, 1.2\%, and 2.0\% respectively. Moreover, when combined with other data augmentation methods - G\textsc{rand}, we can still improve by a margin of 0.3\%, 0.4\%, and 0.7\% respectively. Furthermore, based on the std information of the experimental results of G\textsc{rand} and our LA-G\textsc{rand}, we compute the p-value by t-test to verify the improvements. Except for LA-GCN v.s. G\textsc{rand}-GCN on Cora (with p-value 0.046), all the p-value $\ll$ 0.01 by t-test (the same test is also employed by GRAND), which shows the improvements of LA- over G\textsc{rand}- are statistically significant. Compared with other data augmentation models~\citep{zhu2020pre, rong2020dropedge, zhao2021data}, LA-GNN achieves the best performance on two popular backbones GCN and GAT, showing local information is indeed better than the augmentation approaches from a global perspective, such as DropEdge~\citep{rong2020dropedge} and G\textsc{aug}~\citep{zhao2021data}. Both our model and GraphSNN start from the perspective of the subgraph. Results show that local augmentation is more effective than GraphSNN in capturing the feature information of the local neighborhood, which demonstrates that it's better to consider feature and structure information in designing subgraph-related GNNs.

\subsection{Full-supervised Learning}
\begin{table}[t]
\setlength{\tabcolsep}{0.9mm}
	\caption{Test performance (\%) averaged over 10 runs on node property prediction.  Blank denotes no statistics on the leaderboard or in the paper.}
	\label{tab:ogbn-results}
	\vskip 0.15in
\centering
\begin{tabular}{lccc}
\hline
 	& \texttt{products} & \texttt{proteins} & \texttt{arxiv}   \\
 Model	& Acc      & ROC-AUC &  Acc \\
\hline\hline
MLP & 61.06{$\pm$0.08} & 72.04{$\pm$0.48}  & 55.50{$\pm$0.23}   \\
CoLinkDistMLP & 62.59{$\pm$0.10} & -& 56.38{$\pm$0.16}    \\
Node2vec & 72.49{$\pm$0.10} & 68.81{$\pm$0.65}  & 70.07{$\pm$0.13}    \\
GraphZoom & 74.06{$\pm$0.26} & -& 71.18{$\pm$0.18}    \\
\hline
GCN & 75.64{$\pm$0.21} & 72.51{$\pm$0.35}  & 71.74{$\pm$0.29}    \\
+FLAG & -  & 71.71{$\pm$0.50} &  72.04{$\pm$0.20}     \\
+GraphSNN & -  & - &  \textbf{72.20{$\pm$0.90}}   \\
+LA & \textbf{76.11{$\pm$0.09}}  & \textbf{73.25{$\pm$0.51}} &  72.08{$\pm$0.14}      \\
\hline
GraphSAGE     & 78.70{$\pm$0.36} & 77.68{$\pm$0.20}  & 71.49{$\pm$0.27}     \\
+FLAG   & 79.36{$\pm$0.57}  & 76.57{$\pm$0.75}  &72.19{$\pm$0.21}           \\
+GraphSNN   & -  & -  &71.80{$\pm$0.70}           \\
+LA   & \textbf{79.44{$\pm$0.25}}  & \textbf{77.86{$\pm$0.37}} &\textbf{72.30{$\pm$0.12}}         \\
\hline
GAT & 79.45{$\pm$0.59}  &   -     &      73.65{$\pm$0.11}         \\
+FLAG    & \textbf{81.76{$\pm$0.45}}  &  -   & 73.71{$\pm$0.13}           \\
+LA    &80.46{$\pm$0.54}   &  -   & \textbf{73.77{$\pm$0.12}}       \\
\hline
	\end{tabular}
\end{table}
\paragraph{Datasets.}
To demonstrate the effectiveness of our model on large graphs for full-supervised node and link classification tasks, we utilize ogbn-products, ogbn-proteins, ogbn-arxiv, and ogbl-collab datasets from Open Graph Benchmark (OGB)~\citep{hu2020open} for evaluation. All the dataset statistics can be found in the Appendix~\ref{appendix:reproducibility}.
\paragraph{Baselines.}
We consider four popular message-passing GNNs: GCN~\citep{kipf2017semi}, GAT~\citep{velivckovic2018graph}, and GraphSAGE~\citep{hamilton2017inductive} as backbones. For each of these backbones, we apply the concatenation-style or average-style design discussed in Sec.~\ref{architecture} as our LAGNN architecture, which is detailed in the Appendix~\ref{appendix:reproducibility}. For node classification on arxiv, proteins, and products, we compare it against MLP, Node2vec~\citep{grover2016node2vec}, GCN, GAT, GraphSAGE, FLAG~\citep{kong2020flag}, GraphSNN~\citep{wijesinghe2022a}, GraphZoom~\citep{deng2020graphzoom}, and CoLinkDistMLP~\citep{luo2021distilling}. Besides, we use ogbl-collab to evaluate the performance of our model on the link prediction task, and compare it against MLP, Node2vec, GCN, GraphSAGE. 
\paragraph{Experimental Setup and Results.}
We follow the experimental setup as in OGB~\citep{hu2020open}. For the detailed setup, such as the split ratio and evaluation metric, we just follow the same setting from the OGB implementation. Note that the test results of the baselines are from the official OGB leaderboard (\url{https://ogb.stanford.edu/}) or corresponding papers. For a fair comparison, we implement our models on OGB tasks from the open-resource codes with only touching the first layer. From the OGB leaderboard, we can know that the test results are sensitive concerning model size and various tricks. So we do not change the model size of the backbones as suggested in Sec.~\ref{architecture} and do not add other tricks. Results are summarized in Table~\ref{tab:ogbn-results} and Table~\ref{tab:ogbl-results}. Following common practice, we report the test accuracy associated with the best validation accuracy. The results on node and link prediction demonstrate that our augmentation model consistently improves performance over backbones.
\begin{table}[t]
\setlength{\tabcolsep}{5mm}
\caption{Test performance (\%) averaged over 10 runs on link prediction.}
	\label{tab:ogbl-results}
	\vskip 0.15in
        \centering
        \begin{tabular}{lc}
            \hline
            & \texttt{ogbl-collab} \\
            Model	& Hits@50 (\%)  \\
            \hline\hline
            MLP & 19.27{$\pm$1.29} \\
            Node2vec & 48.88{$\pm$0.54} \\
            \hline
             GCN  & 44.75{$\pm$1.07} \\
            +LA & \textbf{47.49{$\pm$1.40}} \\
            \hline
            GraphSAGE & 48.10{$\pm$0.81} \\
            +LA & \textbf{49.23{$\pm$0.55}} \\
            \hline
        \end{tabular}
\end{table}

\begin{table}[t]
\setlength{\tabcolsep}{2mm}
\caption{Test performance (\%) averaged over 10 runs on graph property prediction.}
	\label{tab:ogbg-results}
	\vskip 0.15in
        \centering
        \begin{tabular}{lcc}
            \hline
            & \texttt{ogbg-molhiv} & \texttt{ogbg-molpcba} \\
            Model	& ROC-AUC & AP  \\
            \hline\hline
             GCN  & 76.06{$\pm$0.97} & 20.20{$\pm$0.24} \\
            +LA & \textbf{76.18{$\pm$1.11}}  & \textbf{{20.28$\pm$0.16}} \\
            \hline
            GIN & \textbf{75.58{$\pm$1.40}} & \textbf{22.66{$\pm$0.28}}\\
            +LA & 75.20{$\pm$}1.74 & {22.38$\pm$0.24} \\
            \hline
        \end{tabular}
\end{table}

\subsection{Inductive Learning}
To evaluate the effectiveness of our model on inductive learning tasks, we take ogbg-molhiv and ogbg-molpcba datasets from OGB for evaluation. For the experimental setup, we just follow the official OGB implementation. We consider GCN and GIN~\citep{xu2019powerful} as backbones. The results are summarized in Table~\ref{tab:ogbg-results}. The experimental results show that our model still works for GCN on inductive learning tasks. Our generative model is only trained on the training dataset. As long as the graphs on the test dataset and the training dataset have similar distributions, $i.e.$, similar subgraph structures and feature vectors, our generative model can make reasonable inferences and generate effective augmented feature vectors.
\subsection{Ablation Study}
\begin{table}[t]
\setlength{\tabcolsep}{1.5mm}
\caption{Effects of different components of our LAGCN evaluated on Pubmed dataset.}
\label{tab:ablation}
\vskip 0.15in
    \centering
    \begin{tabular}{l|c| c | c}
    \toprule
    Technique & Accuracy (\%) &  $\Delta$ & Cumu $\Delta$\\
    \midrule
    GCN & 79.0 & 0 & 0 \\
    + Concatenation & 79.3{$\pm$0.4} & 0.3 &0.3 \\
    + Local Augmentation & 81.1{$\pm$0.5} & 1.8 &2.1 \\
    + Consistency Training & 81.4{$\pm$0.5}  &0.3  &2.4 \\
    + Sharpening Trick & 81.7{$\pm$0.7} &0.3  &2.7 \\
    \bottomrule
    \end{tabular}
\end{table}
\label{Ablation Study}
To demonstrate the effectiveness of our proposed local augmentation model, we conduct experiments of LAGCN on Pubmed that compare it to several of its ablated variants. The results are shown in Table~\ref{tab:ablation}. ``+ concatenation" means that we only apply our concatenation-style design architecture of LAGCN in Sec.~\ref{lagcn} with the original feature matrix as additional concatenated input. The improvement is 0.3\%, which shows that our modification of architecture does not have a lot of effect on the result. ``+ local augmentation" means we use the generated feature matrix as additional concatenated input without consistency training. Although we do not use consistency training, the generated feature matrix as additional input improves the GCN's test accuracy by a margin of 1.8\%. With the consistency training and sharpening trick, we can enhance the performance further. From the ablation study, it is evident that the performance improvement is mainly due to our local augmentation.

\begin{table}[t]
\setlength{\tabcolsep}{0.7mm}
\caption{Summary of results of GCN evaluated on Citeseer on recovering study in terms of classification accuracy (\%). $\downarrow$ means a decrease compared with the accuracy if features are not masked.}
\label{tab:missing}
\vskip 0.15in
    \centering
    \begin{tabular}{l|c| c | c | c}
    \toprule
    Mask Ratio & 0.1& 0.2 & 0.4 & 0.8\\
    \midrule
    GCN & 70.4($\uparrow$0.1)  & 69.2($\downarrow$1.1)  & 67.2($\downarrow$3.1)  &61.1($\downarrow$9.2)  \\
    LAGCN &73.8($\downarrow$0.9)   & 74.0($\downarrow$0.7)  & 71.8($\downarrow$2.9)  & 68.7($\downarrow$6.0)\\
    \bottomrule
    \end{tabular}
\end{table}
\subsection{Robustness to Missing Information}
In this section, we conduct experiments to verify that our proposed framework is robust against missing information in the feature attributes. Specifically, we mask a certain percentage of the attributes of each feature vector and use the same pipeline to do augmentation for the masked feature matrix. As shown in Table~\ref{tab:missing}, we can see that as the mask ratio increases, the gap of the performance between the GCN and LA-GCN enlarges in most cases in Citeseer, which corroborates our insight that our local augmentation can complement the contextual information of the local neighborhood.

\begin{table}[t]
\setlength{\tabcolsep}{5mm}
\caption{Summary of results of GCN and LAGCN evaluated on Pubmed on case study in terms of classification accuracy (\%). Node degree is computed on $\widetilde{\bm{D}}^{-\frac{1}{2}} \widetilde{\bm{A}} \widetilde{\bm{D}}^{-\frac{1}{2}}$.}
\label{tab:case study}
\vskip 0.15in
    \centering
    \begin{tabular}{l|c| c}
    \toprule
    Degree & $[2, 5]$ & $[6, 20]$\\
    \midrule
    \#Nodes & 761 &189 \\
    GCN & 78.2 &82.0 \\
    LAGCN &79.9 &82.2  \\
    $\Delta$ &1.7 &0.2 \\
    \bottomrule
    \end{tabular}
\end{table}
\begin{table}[t]
\setlength{\tabcolsep}{3mm}
\caption{The MADgap metric of LAGCN v.s. GCN of 10 runs on Cora (on different layers).}
\label{tab:madgap}
\vskip 0.15in
    \centering
    \begin{tabular}{l|c| c}
    \toprule
    Layer & GCN &  LAGCN\\
    \midrule
    Layer2 & 0.63{$\pm$0.02} & 0.68{$\pm$0.02} \\
    Layer3 & 0.61{$\pm$0.08} & 0.61{$\pm$0.07} \\
    Layer4 & 0.55{$\pm$0.06} & 0.64{$\pm$0.05} \\
    Layer5 &0.39{$\pm$0.23}  &0.61{$\pm$0.06} \\
    Layer6 &0.24{$\pm$0.47}  &0.31{$\pm$0.17} \\
    \bottomrule
    \end{tabular}
\end{table}
\subsection{Case Study}
In this section, we explore the change in test accuracy of different nodes after applying our local augmentation method. Note that we only apply local augmentation without consistency training, and we set $K$ to 1. From Table~\ref{tab:case study}, we can draw the following conclusions: 1) The degree of most nodes on the Pubmed test set is relatively small, of which with degree fewer than 6 accounts for about 76.1\%. 2) Nodes with smaller degree tend to have lower test accuracy. However, our local augmentation can enrich local information for these nodes and thus enhance their performance. 

\subsection{Over-smoothing Analysis}
It is well known that stacking GNN layers leads to over-smoothing~\citep{li2018deeper}. In this section, we discuss how our proposed approach prevents the over-smoothing issue in GNNs compared to existing approaches. We utilize the MADgap~\citep{chen2020measuring} metric to compare our method with existing GNNs. Table~\ref{tab:madgap} reports the MADgap metric of LAGCN and GCN on Cora (on different layers). We can observe that the MADgap metric of LAGCN is larger than or the same as that of GCN on different layers. Although our approach is not to address over-smoothing, our method can enrich the local neighborhood information and thus can improve the locality of node representations. Therefore, we can alleviate over-smoothing.

\section{Related Work}
\paragraph{Unsupervised Representation Learning on Graphs.} 
In general, unsupervised representation learning methods on graphs include contrastive-based self-supervision methods~\citep{velickovic2019deep,sun2020infograph,hassani2020contrastive,you2020graph}, graph embedding methods~\citep{garcia2017learning, hamilton2017inductive}, and random walk methods~\citep{perozzi2014deepwalk,tang2015line,grover2016node2vec}. Contrastive learning works~\citep{hassani2020contrastive,you2020graph} employ contrastive loss functions to minimize the representation distance of the positive pairs and maximize the distance of the negative pairs. Random walk methods get sentences by taking random walks across nodes and use NLP word embedding models to learn node representations. Our local augmentation is also an unsupervised method for learning local neighborhood information.

\paragraph{Graph Generative Models.} 
Generative models~\citep{goodfellow2014generative, kingma2013auto} are powerful tools of learning data distribution. Recently, researchers have proposed several interesting generative models for graph data generation. Variational graph auto-encoder (VGAE)~\citep{kipf2016variational} exploits the latent variables to learn interpretable representations for undirected graphs. \citet{salha2019keep} make use of a simple linear model to replace the GCN encoder in VGAE and reduce the complexity of encoding schemes. \citet{xu2019generative} propose a generative GCN model to learn node representations for growing graphs. ConDgen~\citep{yang2019conditional} exploits the GCN encoder to handle the invariant permutation for conditional structure generation. Besides, some methods have been proposed to apply the graph generative models in various applications such as graph matching~\citep{simonovsky2018graphvae}, and molecule design~\citep{liu2018constrained}, retrosynthesis prediction~\citep{shi2020graph} and chemical design~\citep{samanta2018designing}. Compared with these approaches mainly focusing on structure generation, our model takes full use of the power of the generative model for feature representation generation, which can serve as an enhanced technique for the downstream backbone models.

\paragraph{Concatenation-style Design.}
In this work, we use concatenation to concatenate the original features and the different generated features at each training iteration to enrich the neighborhood information through local augmentation. Concatenation-style design is a general technique that many works employ such as GAT~\citep{velivckovic2018graph} and SIGN~\citep{rossi2020sign}. SIGN focuses on the scalable training of GNN models on large graph and augments the feature by multi-hop information through powers of adjacency.
 
\section{Conclusion}
We propose local augmentation, a new technique that exploits the generative model to learn the conditional distribution of the central node's neighbors' features given the central node's feature. We feed the generated feature matrix from a well-trained generative model to some modified backbone GNN models to enhance their performance. Experiments show that our model can improve performance across various GNN architectures and benchmark datasets. Besides, our model achieves new state-of-the-art results on various semi-supervised node classification tasks. One limitation of our proposed framework is that we do not exploit the 2-hop neighbors or use the random walk to find more related neighbors for the central node. The future work is to extract more 2/3-hop neighbors if the central node's degree is small and learn the conditional distribution for random sampling nodes if the graph is large.
\section*{Acknowledgements}
We thank all the reviewers for their helpful comments and suggestions, the support of Tencent AI Lab and the helpful discussion from Yulai Cong. Specially, Songtao Liu is also thankful for the encouragement from Hao Yin. Liu and Wu were supported partially by an NSF grant CNS-1652790 and a seed grant from Penn State Center for Security Research and Education (CSRE).


\bibliography{example_paper}
\bibliographystyle{icml2022}

\newpage
\appendix
\onecolumn
\section{Proofs}
We give more details of the derivation of the generator ELBO (Eq. (\ref{generator elbo})) as follows:
\begin{equation*}
\begin{split}
	\log p_{\theta}(\bm{X}_u|\bm{X}_v)&= 
	\int q_{\phi}(\mathbf{z}|\bm{X}_u,\bm{X}_v)\log p_{\theta}(\bm{X}_v|\bm{X}_v)\mathrm{d}\mathbf{z} \\
	&=\int q_{\phi}(\mathbf{z}|\bm{X}_u,\bm{X}_v)\log \frac{p_{\theta}(\bm{X}_u,\bm{X}_v)}{p_{\theta}(\bm{X}_v)}\mathrm{d}\mathbf{z} \\
	&=\int q_{\phi}(\mathbf{z}|\bm{X}_u,\bm{X}_v)\log \frac{p_{\theta}(\bm{X}_u,\bm{X}_v)p_{\theta}(\bm{X}_v,\bm{X}_v,\mathbf{z})}{p_{\theta}(\bm{X}_v)p_{\theta}(\bm{X}_v,\bm{X}_v,\mathbf{z})}\mathrm{d}\mathbf{z} \\
	&=\int q_{\phi}(\mathbf{z}|\bm{X}_v,\bm{X}_v)\log \frac{p_{\theta}(\bm{X}_u,\bm{X}_v,\mathbf{z})}{p_{\theta}(\bm{X}_v)}\frac{1}{\frac{p_{\theta}(\bm{X}_u,\bm{X}_v,\mathbf{z})}{p_{\theta}(\bm{X}_u,\bm{X}_v)}}\mathrm{d}\mathbf{z} \\
	&=\int q_{\phi}(\mathbf{z}|\bm{X}_u,\bm{X}_v)\log \frac{p_{\theta}(\bm{X}_u,\mathbf{z}|\bm{X}_v)}{p_{\theta}(\mathbf{z}|\bm{X}_u,\bm{X}_v)}\mathrm{d}\mathbf{z} \\
	&=\int q_{\phi}(\mathbf{z}|\bm{X}_u,\bm{X}_v)\log \frac{p_{\theta}(\bm{X}_u,\mathbf{z}|\bm{X}_v)}{p_{\theta}(\mathbf{z}|\bm{X}_u,\bm{X}_v)}\frac{q_{\phi}(\mathbf{z}|\bm{X}_u,\bm{X}_v)}{q_{\phi}(\mathbf{z}|\bm{X}_u,\bm{X}_v)}\mathrm{d}\mathbf{z} \\
	&=\int q_{\phi}(\mathbf{z}|\bm{X}_u,\bm{X}_v)\left(\log \frac{p_{\theta}(\bm{X}_u,\mathbf{z}|\bm{X}_v)}{q_{\phi}(\mathbf{z}|\bm{X}_u,\bm{X}_v)}+\log \frac{q_{\phi}(\mathbf{z}|\bm{X}_u,\bm{X}_v)}{p_{\theta}(\mathbf{z}|\bm{X}_u,\bm{X}_v)}\right)\mathrm{d}\mathbf{z} \\
	&=\int q_{\phi}(\mathbf{z}|\bm{X}_u,\bm{X}_v)\log \frac{p_{\theta}(\bm{X}_u,\mathbf{z}|\bm{X}_v)}{q_{\phi}(\mathbf{z}|\bm{X}_u,\bm{X}_v)}\mathrm{d}\mathbf{z} + KL(q_{\phi}(\mathbf{z}|\bm{X}_u, \bm{X}_v)|| p_{\theta}(\mathbf{z}|\bm{X}_u, \bm{X}_v))\\
	&\geq \int q_{\phi}(\mathbf{z}|\bm{X}_u,\bm{X}_v)\log \frac{p_{\theta}(\bm{X}_u,\mathbf{z}|\bm{X}_v)}{q_{\phi}(\mathbf{z}|\bm{X}_u,\bm{X}_v)}\mathrm{d}\mathbf{z}
\end{split}
\end{equation*}

\begin{equation*}
\begin{split}
L_{ELBO} &= \int q_{\phi}(\mathbf{z}|\bm{X}_u,\bm{X}_v)\log \frac{p_{\theta}(\bm{X}_u,\mathbf{z}|\bm{X}_v)}{q_{\phi}(\mathbf{z}|\bm{X}_u,\bm{X}_v)}\mathrm{d}\mathbf{z}\\
&=\int q_{\phi}(\mathbf{z}|\bm{X}_u,\bm{X}_v)\log \frac{p_{\theta}(\bm{X}_u,\bm{X}_v,\mathbf{z})}{q_{\phi}(\mathbf{z}|\bm{X}_u,\bm{X}_v)p_{\theta}(\bm{X}_v)}\mathrm{d}\mathbf{z}\\
&=\int q_{\phi}(\mathbf{z}|\bm{X}_u,\bm{X}_v)\log \frac{p_{\theta}(\bm{X}_u|\bm{X}_v,\mathbf{z})p_{\theta}(\bm{X}_v,\mathbf{z})}{q_{\phi}(\mathbf{z}|\bm{X}_u,\bm{X}_v)p_{\theta}(\bm{X}_v)}\mathrm{d}\mathbf{z}\\
&=\int q_{\phi}(\mathbf{z}|\bm{X}_u,\bm{X}_v)\log \frac{p_{\theta}(\bm{X}_u|\bm{X}_v,\mathbf{z})p_{\theta}(\mathbf{z}|\bm{X}_v)}{q_{\phi}(\mathbf{z}|\bm{X}_u,\bm{X}_v)}\mathrm{d}\mathbf{z}\\
&=\int q_{\phi}(\mathbf{z}|\bm{X}_u,\bm{X}_v)\log \frac{p_{\theta}(\mathbf{z}|\bm{X}_v)}{q_{\phi}(\mathbf{z}|\bm{X}_u,\bm{X}_v)}\mathrm{d}\mathbf{z} + \int q_{\phi}(\mathbf{z}|\bm{X}_u,\bm{X}_v)\log p_{\theta}(\bm{X}_u|\bm{X}_v,\mathbf{z})\mathrm{d}\mathbf{z}\\
&=-KL(q_{\phi}(\mathbf{z}|\bm{X}_u,\bm{X}_v)||p_{\theta}(\mathbf{z}|\bm{X}_v))+\int q_{\phi}(\mathbf{z}|\bm{X}_u,\bm{X}_v)\log p_{\theta}(\bm{X}_u|\bm{X}_v,\mathbf{z})\mathrm{d}\mathbf{z} \\
&= -KL(q_{\phi}(\mathbf{z} | \bm{X}_u, \bm{X}_v) \| p_{\theta}(\mathbf{z} | \bm{X}_v))+\frac{1}{L}\sum_{l=1}^L\log p_\theta(\bm{X}_u|\bm{X}_v,\mathbf{z}^{(l)})
\end{split}
\end{equation*}
\section{Pretraining details}
\subsection{Framework}
\begin{figure*}[t]
\centering
\includegraphics[width=12cm,height=5cm]{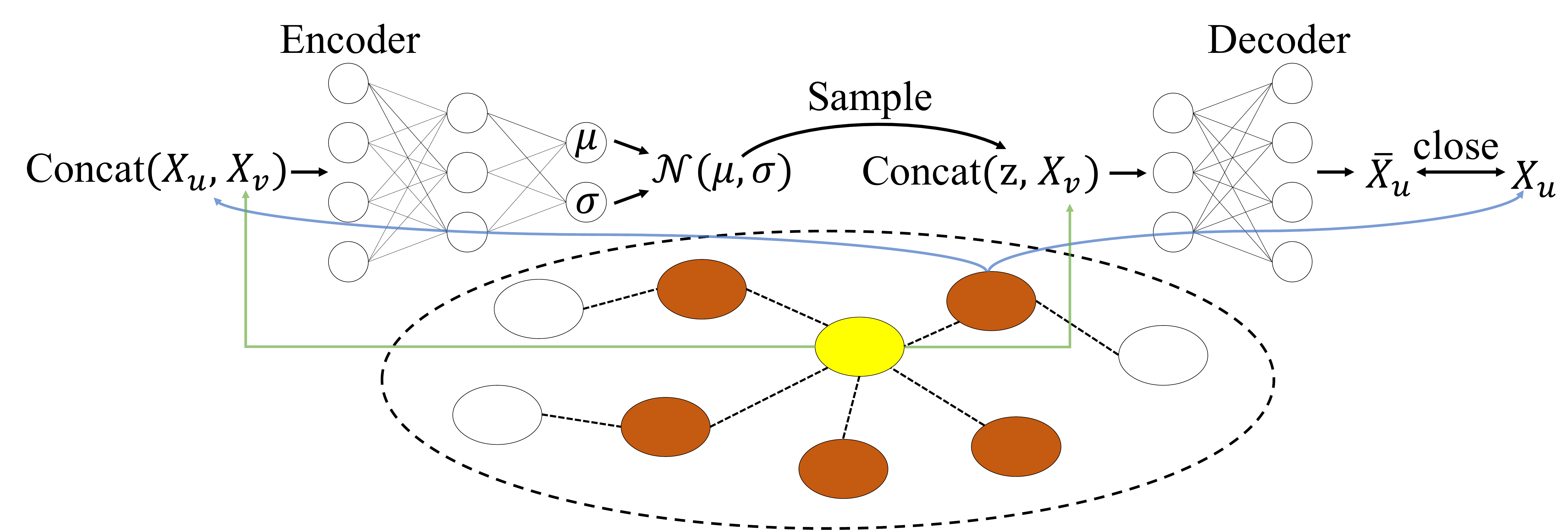}
\caption{A schematic depiction of CVAE. The yellow and brown circles on the graph correspond to the center node and its neighbors respectively. We extract their feature vectors as input for CVAE.}
\label{fig:cvae}
\end{figure*}

We build CVAE based on MLP. The encode and decoder are two-layer MLP where each layer has 256 hidden units for all the graph datasets. For a node $v$ and its neighbors $\mathcal{N}_v$, we extract neighboring-paris ($\bm{X}_v$, $\bm{X}_u$) as input for CVAE during the training stage, where $u\in\mathcal{N}_v$. In the inference stage, we extract a latent variable $\bm{z}$ from $\mathcal{N}(\bm{0}, \bm{I})$ and the center node $v$'s feature vector $\bm{X}_v$ as input for the decoder of CVAE. Thus, we can obtain the generated feature vector $\bm{\overline{X}}_v$. The detail of CVAE is  illustrated in Fig.~\ref{fig:cvae}.

\subsection{Active Learning Trick}
We introduce a trick for the pre-training of CVAE on Cora, Citerseer, and Pubmed. Since the generator may generate some features from the long tail of the distribution. This critical question makes the generation inefficient. Inspired by \citet{nielsen2019gan}, we introduce active learning to address this issue. During active learning, the probability of each feature is proportional to its uncertainty evaluated by an acquisition function. We adopt the Bayesian Active Learning by Disagreement (BALD) acquisition function~\citep{houlsby2011bayesian} with the approximation from the Monte Carlo (MC) dropout samples, which is defined as follows:
\begin{equation}
\label{uncertainty computation}
    U(\bm{\overline{X}}) \approx H\left[\frac{1}{N} \sum_{n=1}^{N} P\left(\bm{Y} | \bm{\overline{X}}, \boldsymbol{\omega}_{n}\right)\right] -\frac{1}{N} \sum_{n=1}^{N} H\left[P\left(\bm{Y} | \bm{\overline{X}}, \boldsymbol{\omega}_{n}\right)\right],
\end{equation}
where $N$ is the number of MC samples and $\boldsymbol{\omega}_{n}$ are the parameters of the network sampled for the $n$-th MC dropout sample. A high BALD score indicates a network with high uncertainty about the generated feature matrix. So it tends to be selected to improve the GNN model. Finally, the CVAE training procedure is summarized in Algorithm~\ref{cvae framework}.

\begin{algorithm}[t]
\small{
\caption{The framework to obtain the Generator $Q_{\Phi}$ with active learning trick on Cora, Citeseer, and Pubmed} 
\hspace*{0.02in} {\bf Input:} 
Adjacency matrix $\bm{A}$, feature matrix $\bm{X}$ \\
\hspace*{0.02in} {\bf Output:} 
Generator $Q_{\Phi}$
\begin{algorithmic}[1]
\STATE Initialize $U$=-inf, $Q_{\Phi}$, and $Q_{\Phi}'$
\FOR{$i=1$ to the number of pre-training epochs} 
    \STATE Update the parameters of generator $Q_{\Phi}$
    \STATE Generate feature matrix $\bm{\overline{X}}$ using $Q_{\Phi}$
    \STATE Compute $U(\bm{\overline{X}})$ using Eq.(\ref{uncertainty computation}).
        \IF{$U(\bm{\overline{X}})>U$} 
            \STATE $U$ = $U(\bm{\overline{X}})$
            \STATE $Q_{\Phi}'$ = $Q_{\Phi}$
        \ENDIF
\ENDFOR
\STATE $Q_{\Phi}$ = $Q_{\Phi}'$
\STATE {\bf Return:} 
Generator $Q_{\Phi}$
\end{algorithmic}
\label{cvae framework}}
\end{algorithm}

\section{Connection to Existing works}
\begin{figure*}[t]
\centering
{\scalebox{0.9}[0.9]{
\subfigure[The Original Graph]{
\begin{minipage}[b]{0.25\textwidth}
\includegraphics[width=1.0\linewidth]{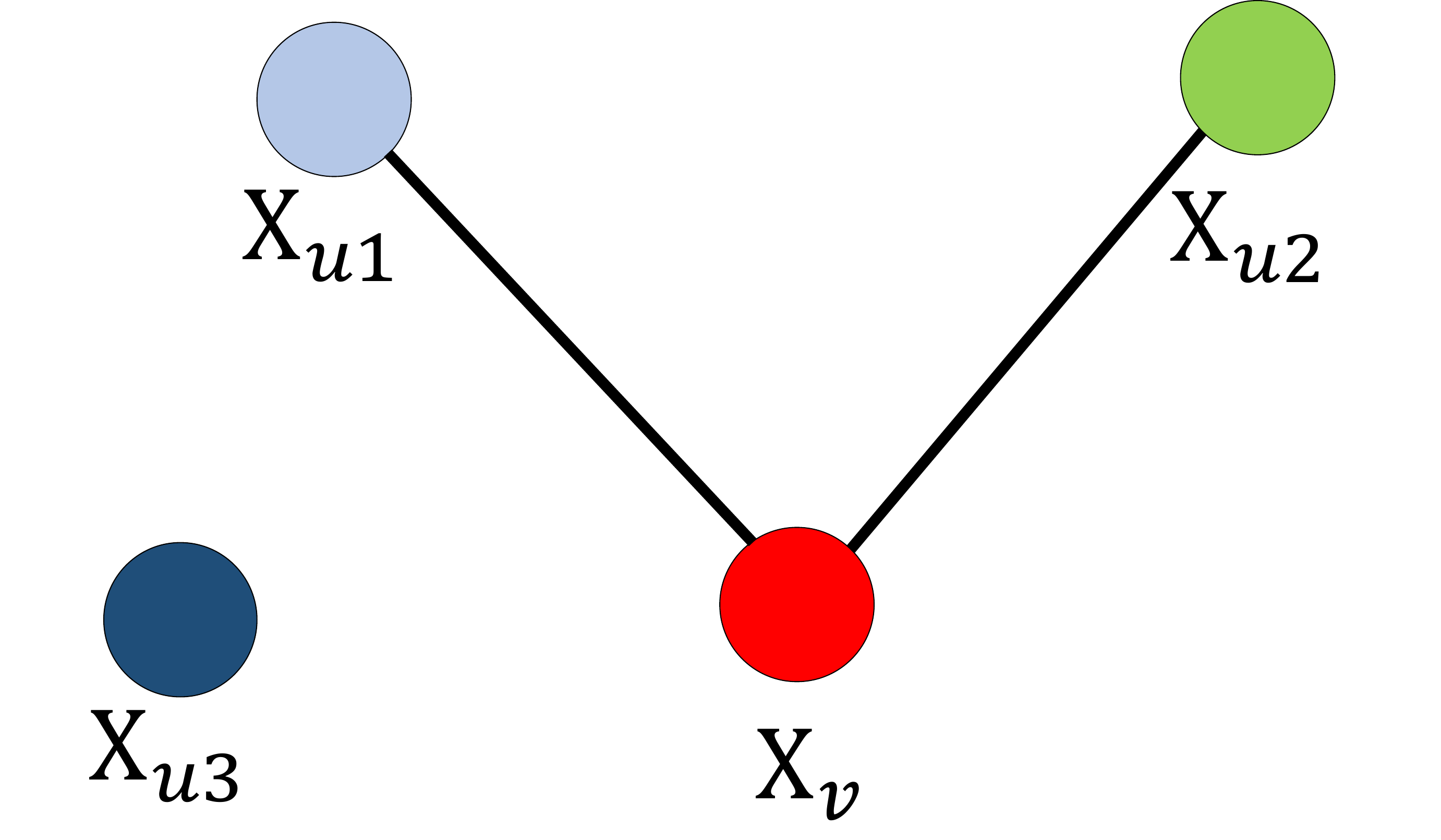}
\end{minipage}}
\subfigure[EP-B]{
\begin{minipage}[b]{0.25\textwidth}
\includegraphics[width=1.0\linewidth]{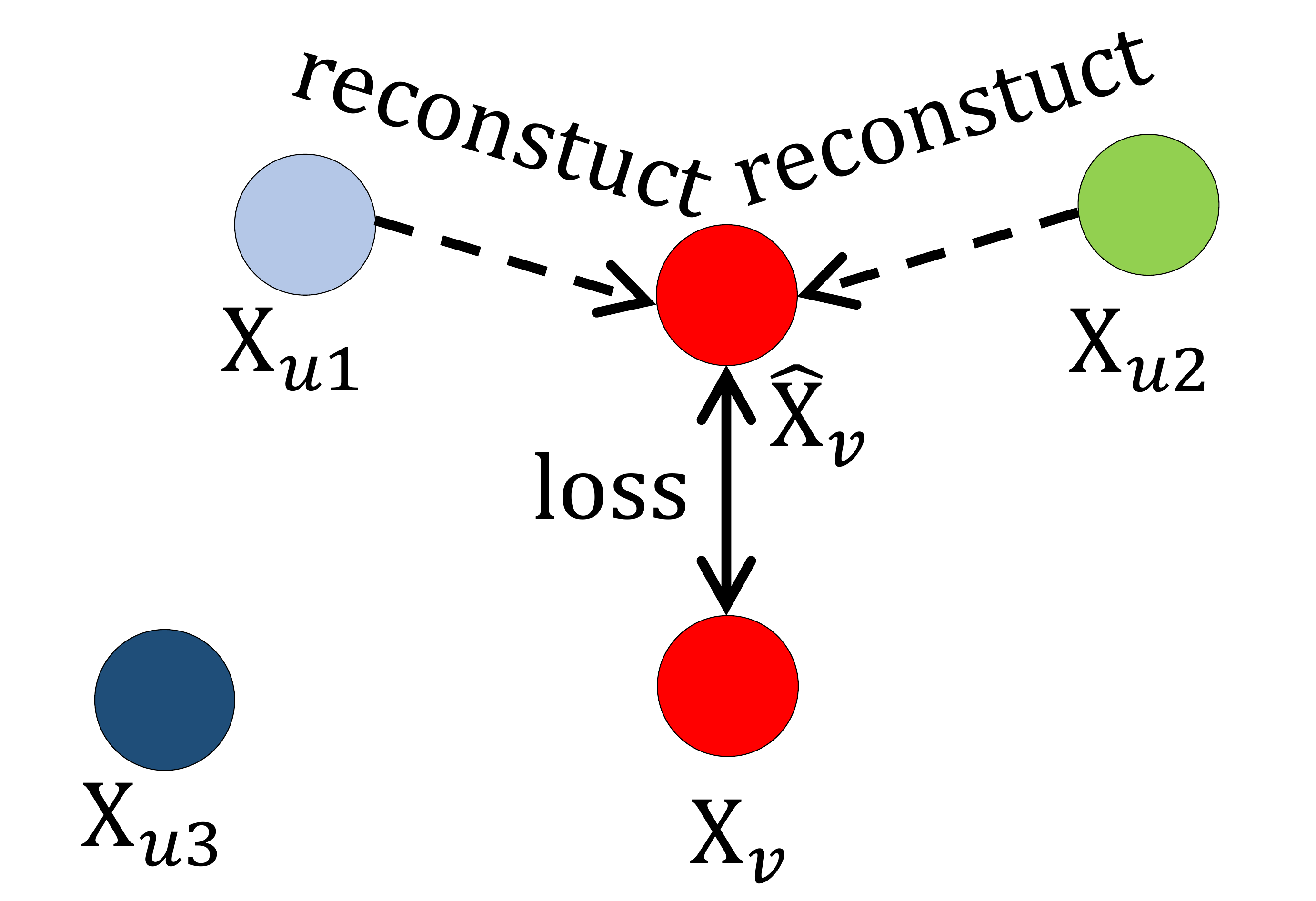}
\end{minipage}}
\subfigure[GraphSAGE]{
\begin{minipage}[b]{0.25\textwidth}
\includegraphics[width=1.0\linewidth]{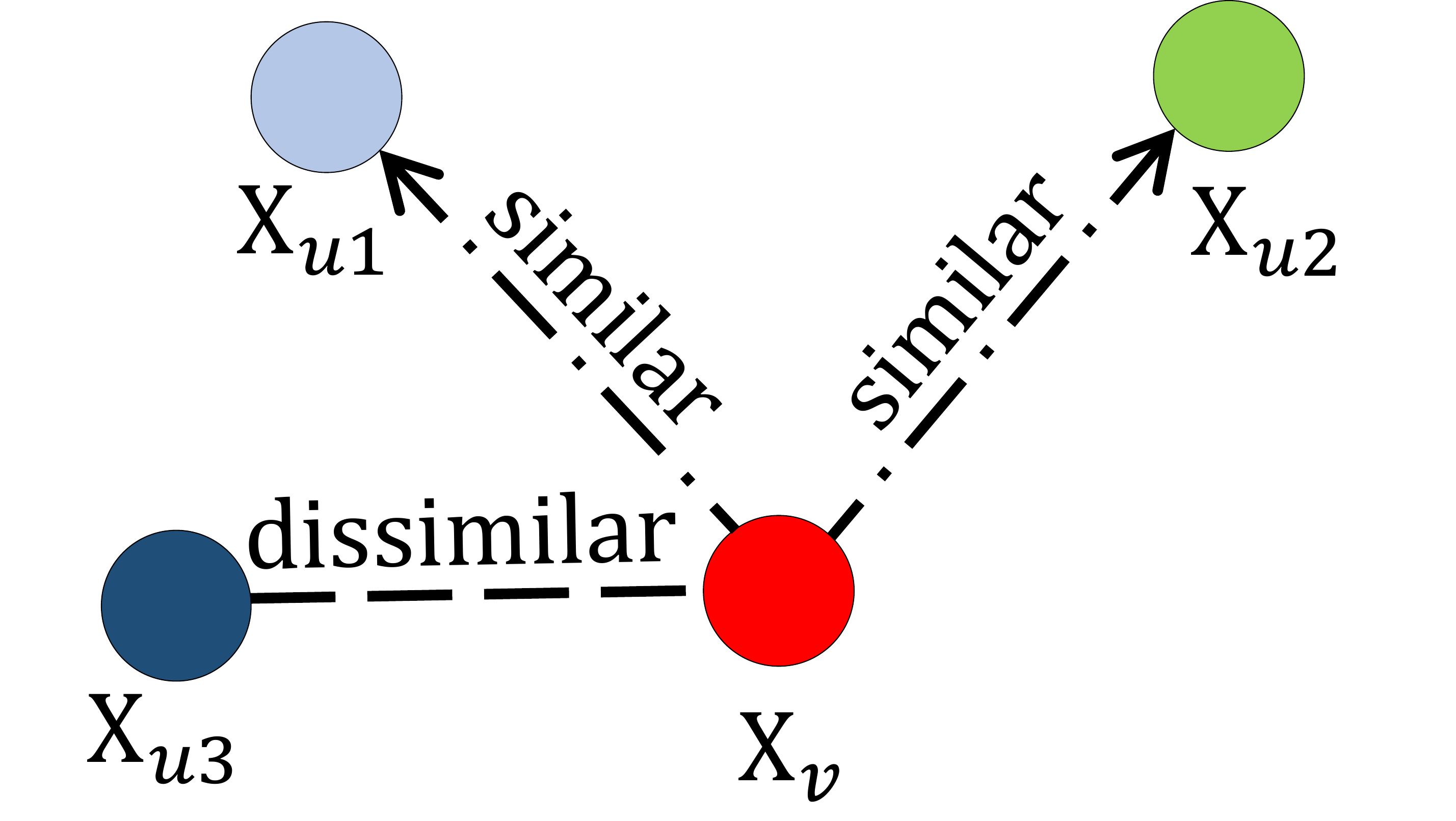}
\end{minipage}}
\subfigure[Local Augmentation]{
\begin{minipage}[b]{0.25\textwidth}
\includegraphics[width=1.0\linewidth]{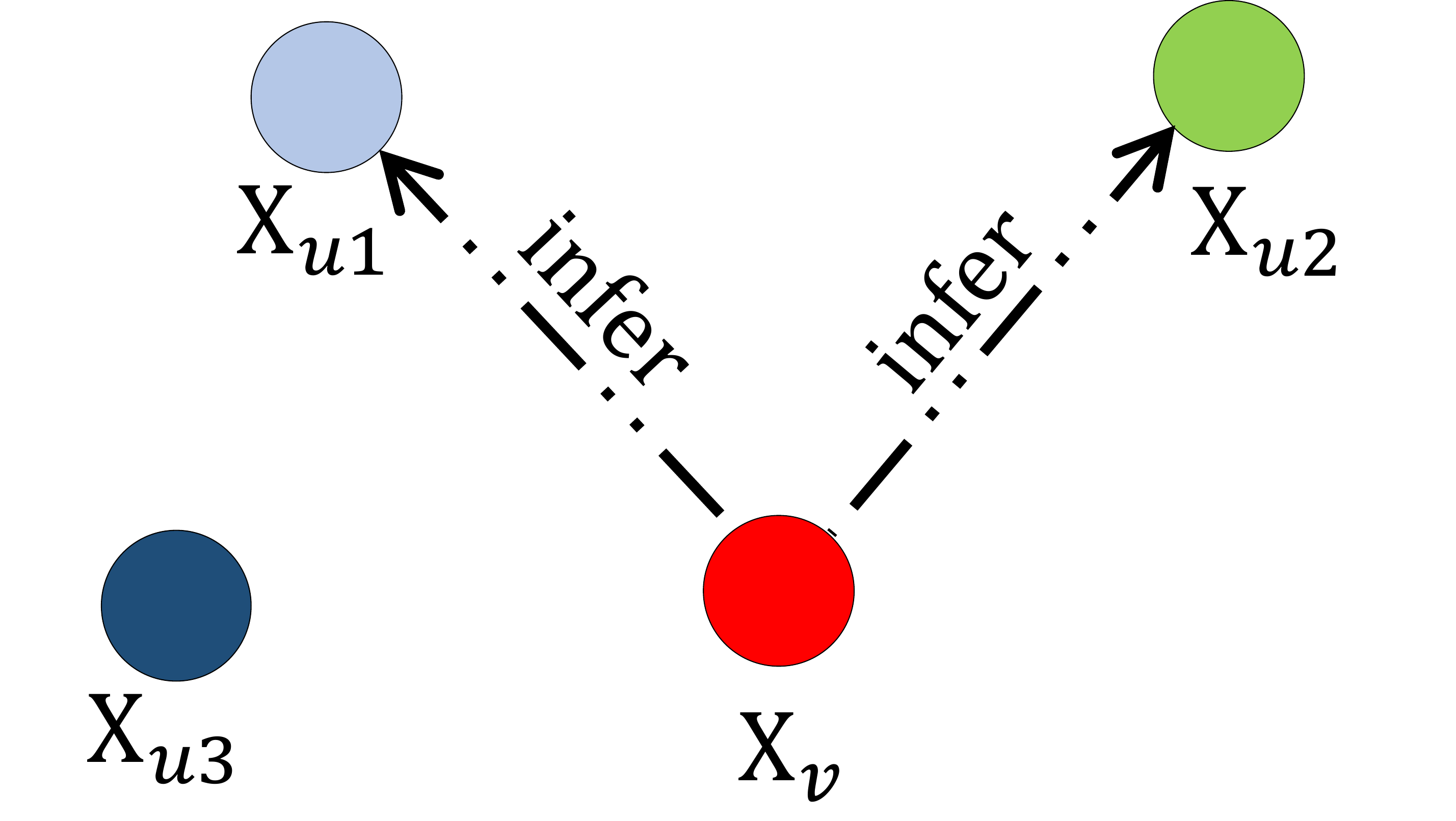}
\end{minipage}}
}}
\caption{(a) The original graph. (b) EP-B exploits the neighbors to reconstruct the central node's embedding. (c) GraphSAGE encourages nearby nodes to have similar embeddings. (d) Given the representation of the central node, our aim is to infer the representations of the connected neighbors.}
\label{fig:compare}
\end{figure*}
\subsection{Connection to EP-B and GraphSAGE}
We discuss how our proposed model distinguishes from the classical graph embedding models. EP-B~\citep{garcia2017learning} and GraphSAGE~\citep{hamilton2017inductive} rely on reconstruction loss function between the center node and its neighbors' embeddings. EP-B aims to minimize their defined reconstruction error and make the attribute vector representation reconstructed by the neighbor node from message passing mechanism is as close as possible to the original attribute vector. GraphSAGE exploits the negative sampling to differentiate the representations of remote node-pairs. GraphSAGE enforce nearby nodes to have similar representations and to enforce disparate nodes to be distinct by minimizing their proposed objective function. The two graph embedding models build upon the assumption that nearby nodes share similar attributes. In contrast, our model does not rely on such assumption and generates more feature vectors of the connected neighbors from a well-learned distribution. A comparison between the reconstruction-based representation learning on graphs and our proposed framework is illustrated in Figure~\ref{fig:compare}. And our local augmentation method is the third paradigm to exploit neighbors in a generative way.

\section{Reproducibility}
\label{appendix:reproducibility}
\begin{table*}[t]
  \centering
  \caption{Data statistics}
  \vskip 0.15in
  \label{tab:stat}
  \renewcommand{\arraystretch}{1.1}
  \resizebox{\linewidth}{!}{
  \begin{tabular}{p{1.6cm}lcccccccc}
    \toprule
      \mr{2}{\textbf{Category}} & \mr{2}{\textbf{Name}} & \mr{2}{\textbf{\#Graphs}} & \textbf{Average} & \textbf{Average} & \mr{2}{\textbf{\#Features}} & \mr{2}{\textbf{\#Classes}} & \textbf{Split} & \textbf{Task} & \mr{2}{\textbf{Metric}} \\
      & & & \textbf{\#Nodes} & \textbf{\#Edges} &  &  & \textbf{Ratio} & \textbf{Type} &\\
    \midrule
    \mr{3}{\shortstack[l]{\textbf{Node}\\$\texttt{citation-}$}}
      & \texttt{cora} & 1 & 2,708 & 5,429 & 1,433 & 7 & 8.5/30.5/61 & Multi-class class. & Accuracy \\
      & \texttt{citeseer} & 1 & 3,327  & 4,732 & 3,703 & 6 & 7.4/30.9/61.7 & Multi-class class. & Accuracy \\
      & \texttt{pubmed} & 1 & 19,717 & 44,338 & 500 & 3 & 3.8/32.1/64.1 & Multi-class class. & Accuracy \\
    \midrule
      \mr{3}{\shortstack[l]{\textbf{Node}\\$\texttt{ogbn-}$}}
      & \texttt{products} & 1 & 2,449,029 & 61,859,140 &100  &47  &8/2/90  & Multi-class class. & Accuracy\\
      & \texttt{proteins} & 1 & 132,534  & 39,561,252 &8 &2  & 65/16/19 & Binary class. & ROC-AUC \\
      & \texttt{arxiv} & 1 & 169,343 & 1,166,243 &128 &40  & 54/18/28 & Multi-class class. & Accuracy \\
    \midrule
      \mr{1}{\shortstack[l]{\textbf{Link}\\$\texttt{ogbl-}$}}
      & \texttt{collab} & 1 & 235,868 & 1,285,465 &128  & - & 92/4/4 & Link prediction & Hits@50 \\
    \midrule
      \mr{2}{\shortstack[l]{\textbf{Graph}\\$\texttt{ogbg-}$}}
      & \texttt{molhiv} & 41,127 & 25.5 & 27.5 &9 &2  & 80/10/10 & Binary class. & ROC-AUC \\
      & \texttt{molpcba} & 437,929 & 26.0 & 28.1 &9  &2  & 80/10/10 & Binary class. & AP \\
    \bottomrule
  \end{tabular}
  }
\end{table*}

\subsection{Datasets Details}
Cora, Citeseer, and Pubmed are standard citation network benchmark datasets~\cite{sen2008collective}. In these datasets, nodes represent documents, and edges denote citations; node features correspond to elements of a bag-of-words representation of a document, and node label corresponds to one of the academic topics. Besides, we utilize six large graph datasets:ogbn-products, ogbn-proteins, ogbn-arxiv, ogbl-collab, ogbg-molhiv, and ogbg-molpcba from OGB~\citep{hu2020open} for evaluation. OGB is a large dataset benchmark used to evaluate GNN models. Specifically, ogbn-products is an co-purchasing network~\citep{Bhatia16} from \citet{chiang2019cluster}. ogbn-arxiv is a citation-type network from \citet{wang2020microsoft}, where nodes represent papers and edges denote citations. ognb-proteins is a protein-protein association network~\citep{szklarczyk2019string}, where nodes represent proteins and edges denote their associations. We use edge embedding as node embedding as suggested in PyG~\citep{fey2019fast} implementation. ogbl-collab is an author collaboration network~\citep{wang2020microsoft}, where nodes represent authors and edges denote their collaboration. The ogbg-molhiv and ogbg-molpcba datasets are from \citet{wu2018moleculenet}, where each graph in the datasets represents a molecule. ogbn-arxiv, ogbn-proteins, ogbn-products are for node classification, ogbl-collab is for link prediction, and ogbg-molhiv and ogbg-molpcba are for graph prediction. All the dataset statistics are summarized in Table~\ref{tab:stat}.

\subsection{Implementation Details}
We use Pytorch~\citep{paszke2019pytorch}, PyG~\citep{fey2019fast}, and DGL~\citep{wang2019dgl} to implement LA-GNNs. The codes of LA-GCN, LA-GAT, LA-GCNII, LA-G\textsc{RAND} on Cora, Citeseer, and Pubmed are implemented referring to Pytorch implementation of GCN\footnote{https://github.com/tkipf/pygcn}~\citep{kipf2017semi}, PyG implementation of GAT\footnote{https://github.com/pyg-team/pytorch\_geometric/blob/master/examples/gat.py}~\citep{velivckovic2018graph}, Pytorch implementation of GCNII\footnote{https://github.com/chennnM/GCNII}~\citep{chen2020simple}, and Pytorch implementation of GRAND\footnote{https://github.com/THUDM/GRAND}~\citep{feng2020grand} respectively. The codes of LA-GCN, LA-GRAND, and LA-GIN on OGB datasets are implemented referring to official OGB implementation\footnote{https://github.com/snap-stanford/ogb/blob/master/examples}. The codes of LA-GAT on ogbn-products and ogbn-arxiv are implemented referring to official OGB implementation\footnote{https://github.com/pyg-team/pytorch\_geometric/blob/master/examples/ogbn\_products\_gat.py} and DGL implementation\footnote{https://github.com/dmlc/dgl/blob/master/examples/pytorch/ogb/ogbn-arxiv/gat.py}. All the experiments in this work are conducted on a single NVIDIA Tesla V100 with 32GB memory size. The software that we use for experiments are Python 3.6.8, pytorch 1.9.0, pytorch-cluster 1.5.9, pytorch-scatter 2.0.9, pytorch-sparse 0.6.12, pyg 2.0.3, ogb 1.3.2, dgl 0.7.2, numpy 1.19.2, torchvision 0.10.0, CUDA 10.2.89, and CUDNN 7.6.5.

\subsection{Hyperparameter Details}
\begin{figure}[t]
\centering
\subfigure[Average-style LAGCN]{
\begin{minipage}[b]{0.49\textwidth}
\centering
\includegraphics[width=0.8\linewidth]{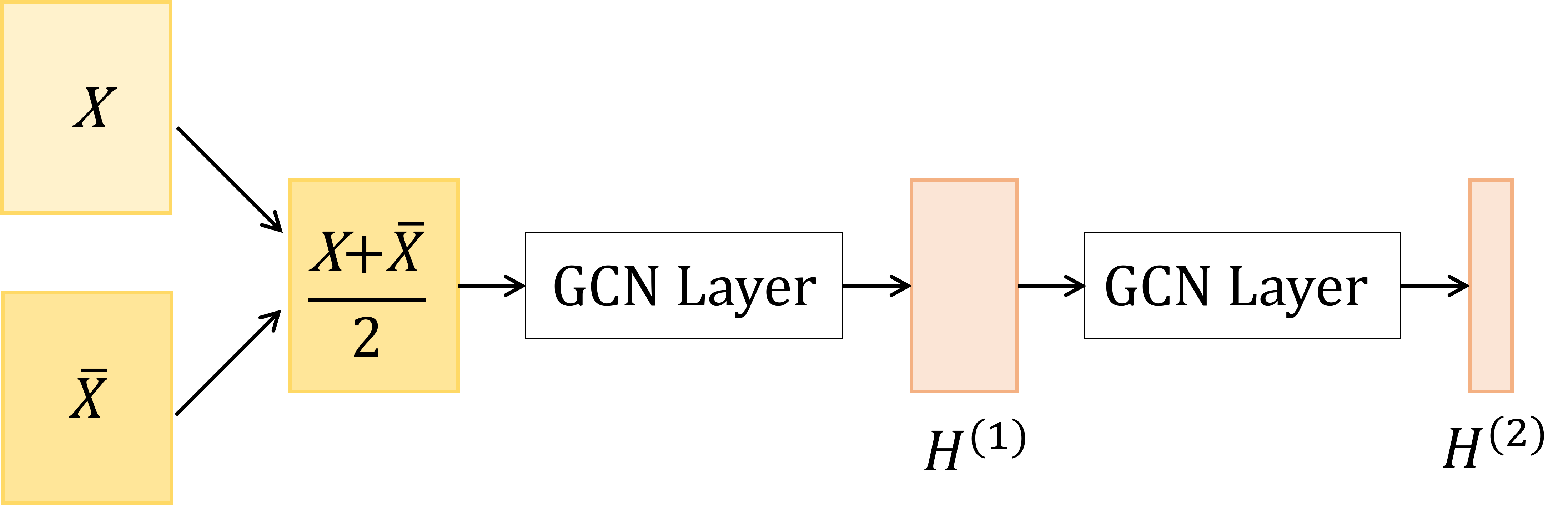}
\end{minipage}}
\subfigure[Concatenation-style LAGCN]{
\begin{minipage}[b]{0.49\textwidth}
\centering
\includegraphics[width=0.8\linewidth]{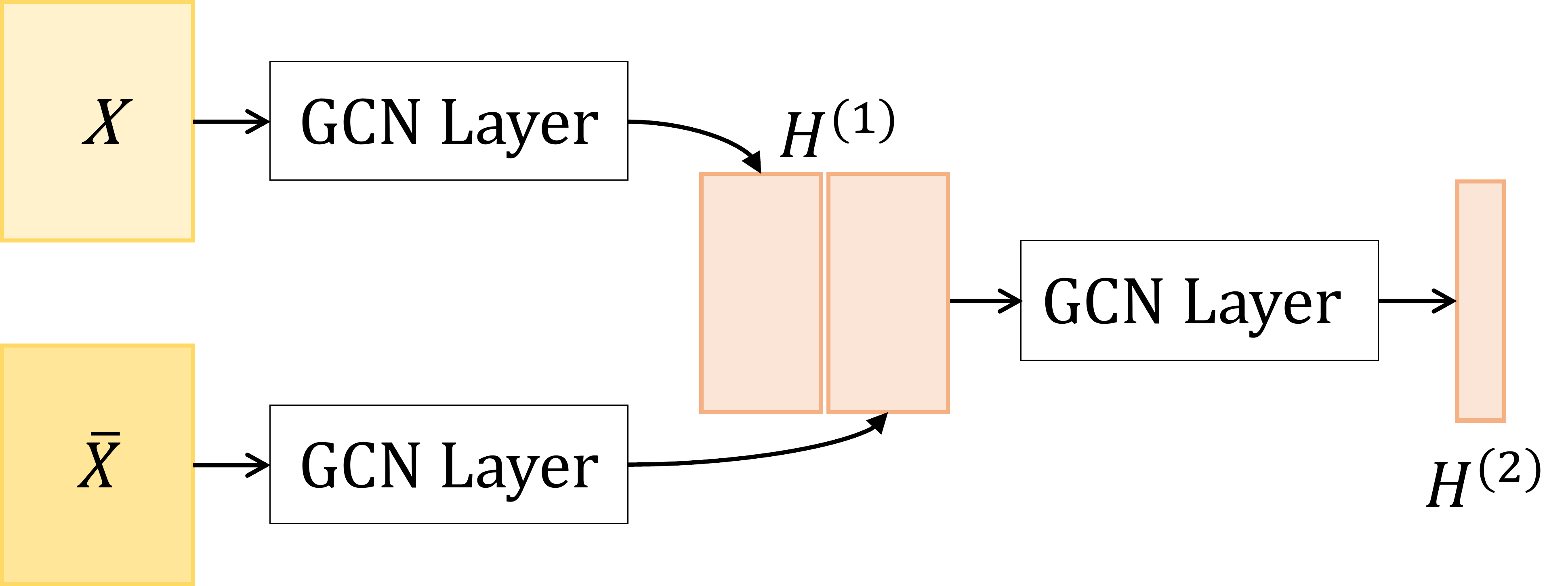}
\end{minipage}}
\caption{Average-style and Concatenation-style LAGCN architectures. The difference between the two architectures is that the concatenation-style LAGCN has an additional convolutional layer for $\bm{\overline{X}}$ and it uses a concatenation operation to mix the hidden representations.}
\label{fig:gcn and la-gcn}
\end{figure}

As we have discussed in Sec.~\ref{architecture}, we provide two design for our LAGNNs - average and concatenation. For concatenation-style design, LA-GNNs introduce an additional aggregation over our generated feature matrix $\bm{\overline{X}}$ before concatenation. The details of two style design architectures of LA-GCN can be found in Figure~\ref{fig:gcn and la-gcn}. Specifically, we use 3 additional generated feature matrix in the first layer for the Pubmed dataset. More details about hyparatemeters can be found in Table~\ref{tab:hyperparameterdetails_citation} and \ref{tab:hyperparameterdetails_ogb}.

\begin{table}[t]
  \centering
  \caption{The hyper-parameters for each backbone on three citation datasets.}
  \label{tab:hyperparameterdetails_citation}
    \vskip 0.15in
    \small
    \begin{tabular}{cl|r|r|p{0.4\textwidth}}
    \hline
    \multicolumn{1}{l}{Dataset} & Backbone & \multicolumn{1}{l|}{Architecture} & \multicolumn{1}{l|}{Additional aggregation} & Hyper-parameters \\
    \hline
    
\multirow{4}[14]{*}{Cora} & LAGCN   & Concatenation     & 1 & epochs: 2000, lr: 0.01, weight-decay: 5e-4, hidden: 8, layers: 2, dropout: 0.5, $K$: 4, consistency training: True, $\beta$: 1.0 $T$: 0.5 \\
    
\cline{2-5} & LAGAT & Concatenation & 1  &epochs: 1000, lr: 0.01, weight-decay: 5e-4, hidden: 8, layers: 2, heads: [4, 1], dropout: 0.6, alpha: 0.2, $K$: 4, consistency training: True, $\beta$: 1.0 $T$: 0.5 \\

\cline{2-5} & LAGCNII & Concatenation  & 1 & lr: 0.01,  $L_{2_{c}}$: 0.01, $L_{2_{d}}$: 5e-4, layers: 64, hidden: 32, $\alpha_l$: 0.1, $\lambda$: 0.5, dropout: 0.6, $K$: 4, consistency training: False, early stopping patience: 200\\

\cline{2-5} & LAG\textsc{rand} & -  & - & lr: 0.01, weight-decay: 5e-4, input dropout rate: 0.5, hidden dropout rate: 0.5, dropNode probability: 0.5, propagation step: 8, hidden: 32, $K$: 4, consistency training: True, $\beta$: 1.0, $T$: 0.5, early stopping patience: 200, batch normalization: False\\
\hline

\multirow{4}[14]{*}{Citeseer} & LAGCN   & Concatenation     & 1 & epochs: 2000, lr: 0.01, weight-decay: 5e-4, hidden: 8, layers: 2, dropout: 0.5, $K$: 4, consistency training: True, $\beta$: 1.0 $T$: 0.5 \\
    
\cline{2-5} & LAGAT & Concatenation & 1  &epochs: 1000, lr: 0.01, weight-decay: 5e-4, hidden: 8, layers: 2, heads: [4, 1], dropout: 0.6, alpha: 0.2, $K$: 4, consistency training: True, $\beta$: 1.0 $T$: 0.5 \\

\cline{2-5} & LAGCNII & Concatenation  & 1 & lr: 0.01,  $L_{2_{c}}$: 0.01, $L_{2_{d}}$: 5e-4, layers: 32, hidden: 128, $\alpha_l$: 0.1, $\lambda$: 0.6, dropout: 0.7, $K$: 4, consistency training: True, $\beta$: 1.0, $T$: 0.5, early stopping patience: 200\\

\cline{2-5} & LAG\textsc{rand} & -  & - & lr: 0.01, weight-decay: 5e-4, input dropout rate: 0.2, hidden dropout rate: 0.1, dropNode probability: 0, propagation step: 2, hidden: 32, $K$: 4, consistency training: True, $\beta$: 0.7, $T$: 0.2, early stopping patience: 200, batch normalization: False \\
\hline
    
\multirow{4}[14]{*}{Pubmed} & LAGCN   & Concatenation     & 3 & epochs: 300, lr: 0.02, weight-decay: 5e-4, hidden: 4, layers: 2, dropout: 0.5, $K$: 4, consistency training: True, $\beta$: 1.0 $T$: 0.5 \\
    
\cline{2-5} & LAGAT & Concatenation & 3  &epochs: 1000, lr: 0.01, weight-decay: 5e-4, hidden: 8, layers: 2, heads: [2, 1], dropout: 0.6, alpha: 0.2, $K$: 4, consistency training: True, $\beta$: 1.0 $T$: 0.5 \\

\cline{2-5} & LAGCNII & Concatenation  & 3 & lr: 0.01,  $L_{2_{c}}$: 5e-4, $L_{2_{d}}$: 5e-4, layers: 16, hidden: 64, $\alpha_l$: 0.1, $\lambda$: 0.4, dropout: 0.5, $K$: 4, consistency training: True, $\beta$: 1.0, $T$: 0.5, early stopping patience: 200\\

\cline{2-5} & LAG\textsc{rand} & -  & - & lr: 0.2, weight-decay: 5e-4, input dropout rate: 0.7, hidden dropout rate: 0.8, dropNode probability: 0.7, propagation step: 5, hidden: 32, $K$: 4, consistency training: True, $\beta$: 1.2, $T$: 0.2, early stopping patience: 200, batch normalization: True \\
\hline
    \end{tabular}
\end{table}

\begin{table}[t]
  \centering
  \caption{The hyper-parameters for each backbone on OGB datasets.}
  \label{tab:hyperparameterdetails_ogb}
    \vskip 0.15in
    \small
    \begin{tabular}{cl|r|r|p{0.4\textwidth}}
    \hline
    \multicolumn{1}{l}{Dataset} & Backbone & \multicolumn{1}{l|}{Architecture} & \multicolumn{1}{l|}{Additional aggregation} & Hyper-parameters \\
    \hline
    
\multirow{3}[9]{*}{ogbn-arxiv} & LAGCN   & Concatenation     & 1 & epochs: 500, lr: 0.01, hidden: 128, layers: 3, dropout: 0.5, $K$: 2, consistency training: True, $\beta$: 1.0 $T$: 0.5 \\
    
\cline{2-5} & LAGAT & Concatenation & 1  &epochs: 2000, lr: 0.01, n-hidden: 125, n-layers: 3, n-heads: 3, dropout: 0.75, mask-rate: 0.5, no-attn-dst: True, use-norm: True, use-labels: True, input-drop: 0, attn-drop: 0, edge-drop: 0, wd: 0, $K$: 2, consistency training: False\\

\cline{2-5} & LASAGE   & Concatenation     & 1 & epochs: 500, lr: 0.01, hidden: 128, layers: 3, dropout: 0.5, $K$: 4, consistency training: True, $\beta$: 1.0 $T$: 0.5 \\
\hline

\multirow{2}[4]{*}{ogbn-proteins} & LAGCN   & Concatenation     & 1 & epochs: 1000, lr: 0.01, hidden: 128, layers: 3, dropout: 0.0, $K$: 2, consistency training: False \\

\cline{2-5} &  LASAGE   & Concatenation     & 1 & epochs: 1000, lr: 0.01, hidden: 128, layers: 3, dropout: 0.0, $K$: 2, consistency training: False \\
\hline
    
\multirow{5}[10]{*}{ogbn-products} & LAGCN   & Average     & - & epochs: 300, lr: 0.01, hidden: 256, layers: 3, dropout: 0.5, $K$: 1, consistency training: False \\
    
\cline{2-5} & LAGAT & Average & -  &epochs: 100, lr: 0.001, hidden: 128, layers: 3, heads: 4, $K$: 2, consistency training: False \\

\cline{2-5} & LASAGE & Average     & - & epochs: 20, lr: 0.001, hidden: 128, layers: 3, dropout: 0.5, heads: 4, $K$: 2, consistency training: True, $\beta$: 1.0 $T$: 0.5 \\
\hline

\multirow{2}[10]{*}{ogbl-collab} & LAGCN   & Concatenation   & 1 & epochs: 400, lr: 0.001, hidden: 128, layers: 3, dropout: 0.0, $K$: 1, consistency training: False \\
    
\cline{2-5} & LASAGE & Concatenation   & 1 & epochs: 400, lr: 0.001, hidden: 128, layers: 3, dropout: 0.0, $K$: 1, consistency training: False \\
\hline

\multirow{2}[10]{*}{ogbg-molhiv} & LAGCN   & Average   & - & epochs: 100, lr: 0.001, hidden: 300, layers: 5, dropout: 0.0, $K$: 1, consistency training: False, batch size: 32 \\
    
\cline{2-5} & LAGIN &  Average   & - & epochs: 100, lr: 0.001, hidden: 300, layers: 5, dropout: 0.0, $K$: 1, consistency training: False, batch size: 32  \\
\hline

\multirow{2}[10]{*}{ogbg-molpcba} & LAGCN   &  Average   & - & epochs: 100, lr: 0.001, hidden: 300, layers: 5, dropout: 0.0, $K$: 1, consistency training: False, batch size: 32  \\
    
\cline{2-5} & LAGIN &  Average   & - & epochs: 100, lr: 0.001, hidden: 300, layers: 5, dropout: 0.0, $K$: 1, consistency training: False, batch size: 32  \\
\hline
    \end{tabular}
\end{table}

\end{document}